\documentclass[misq]{informs3b} 

\OneAndAHalfSpacedXI 

\usepackage{natbib}
\bibpunct[, ]{(}{)}{,}{a}{}{,}%
\def\bibfont{\small}%

\usepackage[resetlabels]{multibib}
\usepackage{booktabs,multirow} 
\newcites{sec}{References}

\TheoremsNumberedThrough     
\ECRepeatTheorems

\EquationsNumberedThrough    

\usepackage{caption}

\usepackage[bottom]{footmisc}
\usepackage{afterpage}

\usepackage{xcolor}

\usepackage[bookmarks, hidelinks]{hyperref}
\hypersetup{
	bookmarksnumbered=true,     
	bookmarksopen=true,
	bookmarksopenlevel=2,    
	colorlinks=true,
	linkcolor={red!50!black},
	citecolor={blue!50!black},
	urlcolor={blue!80!black},
}

\usepackage{bookmark}
\usepackage{etoolbox}
\usepackage{tocloft}
\makeatletter
\pretocmd\endfigure{%
	\addtocontents{lof}{\protect{%
			\bookmark[
			rellevel=1,
			keeplevel,
			dest=\@currentHref,
			]{Figure \thefigure: \@currentlabelname}}}%
	\bookmark[
	rellevel=1,
	keeplevel,
	dest=\@currentHref,
	]{Figure \thefigure: \@currentlabelname}%
}{}{\errmessage{Patching \noexpand\endfigure failed}}
\pretocmd\endtable{%
	\addtocontents{lof}{\protect{%
			\bookmark[
			rellevel=1,
			keeplevel,
			dest=\@currentHref,
			]{Table \thetable: \@currentlabelname}}}%
	\bookmark[
	rellevel=1,
	keeplevel,
	dest=\@currentHref,
	]{Table \thetable: \@currentlabelname}%
}{}{\errmessage{Patching \noexpand\endtable failed}}
\makeatother

\usepackage{tabularx}
\usepackage{cases}
\usepackage{threeparttable}
\usepackage{graphicx}
\usepackage{subcaption}
\emergencystretch=\maxdimen
\makeatletter
\newcommand{\otherlabel}[2]{\protected@edef\@currentlabel{#2}\label{#1}}
\makeatother
\usepackage{dsfont}
\usepackage{ulem}
\usepackage{tablefootnote}
\usepackage{threeparttable}
\usepackage[resetlabels]{multibib}
\newcites{ec}{References} 
\usepackage{blindtext}
\usepackage{lastpage}
\usepackage{fancyhdr}
\usepackage{makecell}

\usepackage{longtable}
\usepackage{caption}
\usepackage{array}
\newcolumntype{P}[1]{>{\raggedright\arraybackslash}p{#1}}

\newcites{ae}{References}
\newcites{rone}{References}
\newcites{rtwo}{References}
\newcites{rthree}{References}

\usepackage{comment}

\usepackage{algorithm} 
\usepackage[noend]{algpseudocode} 
\usepackage{tikz}
\makeatletter
\newcount\dirtree@lvl 
\newcount\dirtree@plvl
\newcount\dirtree@clvl
\def\dirtree@growth{%
	\ifnum\tikznumberofcurrentchild=1\relax
	\global\advance\dirtree@plvl by 1
	\expandafter\xdef\csname dirtree@p@\the\dirtree@plvl\endcsname{\the\dirtree@lvl}
	\fi
	
	\global\advance\dirtree@lvl by 1\relax
	\dirtree@clvl=\dirtree@lvl
	\advance\dirtree@clvl by -\csname dirtree@p@\the\dirtree@plvl\endcsname
	\pgf@xa=1cm\relax
	\pgf@ya=-0.6cm\relax
	\pgf@ya=\dirtree@clvl\pgf@ya
	\pgftransformshift{\pgfqpoint{\the\pgf@xa}{\the\pgf@ya}}%
	
	\ifnum\tikznumberofcurrentchild=\tikznumberofchildren
	\global\advance\dirtree@plvl by -1
	\fi
}
\tikzset{
	dirtree/.style={
		growth function=\dirtree@growth,
		growth parent anchor=south west,
		parent anchor=south west,
		every child node/.style={anchor=west},
		edge from parent path={([xshift=2ex] \tikzparentnode\tikzparentanchor) 
			|- (\tikzchildnode\tikzchildanchor)},
	}
}
\makeatother

\newcommand\Tstrut{\rule{0pt}{2.6ex}}         

\usepackage[toc]{appendix}

\usepackage[nomessages]{fp}

\usepackage{enumitem}
\usepackage{ulem}

\begin{document}

\newcolumntype{L}[1]{>{\raggedright\arraybackslash}p{#1}}
\newcolumntype{C}[1]{>{\centering\arraybackslash}p{#1}}
\newcolumntype{R}[1]{>{\raggedleft\arraybackslash}p{#1}}



\RUNTITLE{Doctor Prototype}




%


\begin{center}
\textbf{\Large Think as a Doctor: An Interpretable AI Approach
for  ICU Mortality Prediction}    
\end{center}

\hspace{0.1cm}
\begin{center}

Qingwen Li$^{1,**}$, Xiaohang Zhao$^{1,**}$, Xiao Han$^{2,*}$, Hailiang Huang$^{1,*}$, Lanjuan Liu$^{1}$ \\
\vspace{0.2cm}
$^1$ School of Information Management \& Engineering, Shanghai University of Finance and Economics, Shanghai, China \\

$^2$ Key Laboratory of Data Intelligence and Management (Beihang University), Ministry of Industry and Information Technology, School of Economics and Management, Beihang University, Beijing, China \\

\vspace{0.1cm}
$*$ Corresponding Author: Xiao Han, \href{mailto:xh_bh@buaa.edu.cn}{xh\_bh@buaa.edu.cn}; Hailiang Huang, \href{mailto:hlhuang@shufe.edu.cn}{hlhuang@shufe.edu.cn}\\
$**$ Equal Contribution \\
\end{center}
\vspace{0.3cm}

\noindent \textbf{Abstract:} 
Intensive Care Unit (ICU) mortality prediction, which estimates a patient's mortality status at discharge using electronic health records (EHRs) collected early in an ICU admission, is vital in high-stakes critical care. 
For this task, predictive accuracy alone is insufficient; interpretability is equally essential for building clinical trust and meeting regulatory standards, a topic that has attracted significant attention in information system research. Accordingly, an ideal solution should enable intrinsic interpretability and align its reasoning process with three key elements of the ICU decision-making practices: clinical course identification, demographic heterogeneity, and prognostication awareness. However, conventional interpretable healthcare prediction approaches largely focus on demographic heterogeneity, overlooking clinical course identification and prognostication awareness. Recent prototype learning methods address clinical course identification, yet the integration of demographic heterogeneity and prognostication awareness into such frameworks remains underexplored. To address these gaps, we propose ProtoDoctor, a novel ICU mortality prediction framework that delivers intrinsic interpretability while integrating all three elements of the ICU decision-making practices into its reasoning process. Methodologically, ProtoDoctor features two key innovations: the Prognostic Clinical Course Identification module and the Demographic Heterogeneity Recognition module.
The former enables the identification of clinical courses via prototype learning and achieves prognostication awareness using a novel regularization mechanism. The latter models demographic heterogeneity through cohort-specific prototypes and risk adjustments. Extensive empirical evaluations demonstrate that ProtoDoctor outperforms state-of-the-art baselines in predictive accuracy. Human evaluations further confirm that its interpretations are more clinically meaningful, trustworthy, and applicable in ICU practice.

\hspace{0.1cm} 


\hspace{1cm}

\section{Introduction}
Fueled by the widespread adoption of electronic health records (EHRs), Information Systems (IS) researchers have tackled numerous healthcare predictive analytics tasks, ranging from readmission prediction to adverse event prediction using advanced IT solutions,  particularly those based on artificial intelligence (AI) techniques~\citep{xie2021readmission,lin2021first,guo2024explainable}.
One important healthcare prediction task is Intensive Care Unit (ICU) mortality prediction, which uses EHR data from the early hours of an ICU admission to predict whether a critically ill patient will survive the hospital stay. This prediction task addresses a significant healthcare need, as in the United States alone, over five million patients are admitted to ICUs each year, and ICU-related expenditures exceed \$100 billion annually \citep{kannan2023growth}. Given the life-threatening conditions of ICU patients and the significant consumption of medical resources, accurate ICU mortality prediction is essential for multiple stakeholders. For patients and their families, reliable mortality risk assessments enable timely and informed treatment decisions. For clinicians and hospital administrators, objective mortality predictions support data-driven care planning and efficient ICU resource allocation. Indeed, one study found that deploying an accurate ICU risk model reduced length of stay and yielded approximately \$1091 in savings per patient \citep{ericson2022potential}. At the macro level, certain AI-based tools could generate substantial financial benefits for the healthcare system, with estimated annual savings ranging from \$55 million to \$72 million \citep{na2023patient}.

Given the importance of the task, numerous methods have been developed for ICU mortality prediction. 
Traditional approaches rely on scoring systems that quantify abnormalities in patients' physiological variables recorded in EHRs. Popular examples include the Simplified Acute Physiology Score (SAPS)~\citep{le1984simplified}, the Acute Physiology and Chronic Health Evaluation (APACHE)~\citep{zimmerman2006acute}, and the Sequential Organ Failure Assessment (SOFA)~\citep{de1996s0fa}. Although easy to interpret, these systems have difficulty in capturing complex and evolving temporal patterns in EHRs, and therefore suffer from limited predictive accuracy and clinical utility ~\citep{salluh2014icu}. In recent years, deep learning (DL) techniques have substantially improved predictive performance by modeling intricate temporal patterns in EHR data \citep{morid2023time}. These methods use architectures such as recurrent, convolutional, and attention-based neural networks to integrate dynamic physiological measurements with static demographic information \citep{harutyunyan2019multitask, caicedo2019iseeu, ma2020concare}. However, these performance gains often come at the cost of interpretability: complex DL models generally function as ``black boxes", which undermines clinician trust and fails to comply with emerging regulatory requirements for algorithmic transparency, e.g., the European Union (EU)'s General Data Protection Regulation (GDPR)~\citep{voigt2017eu}. 

\begin{figure}
    \centering
    \includegraphics[width=0.9\linewidth]{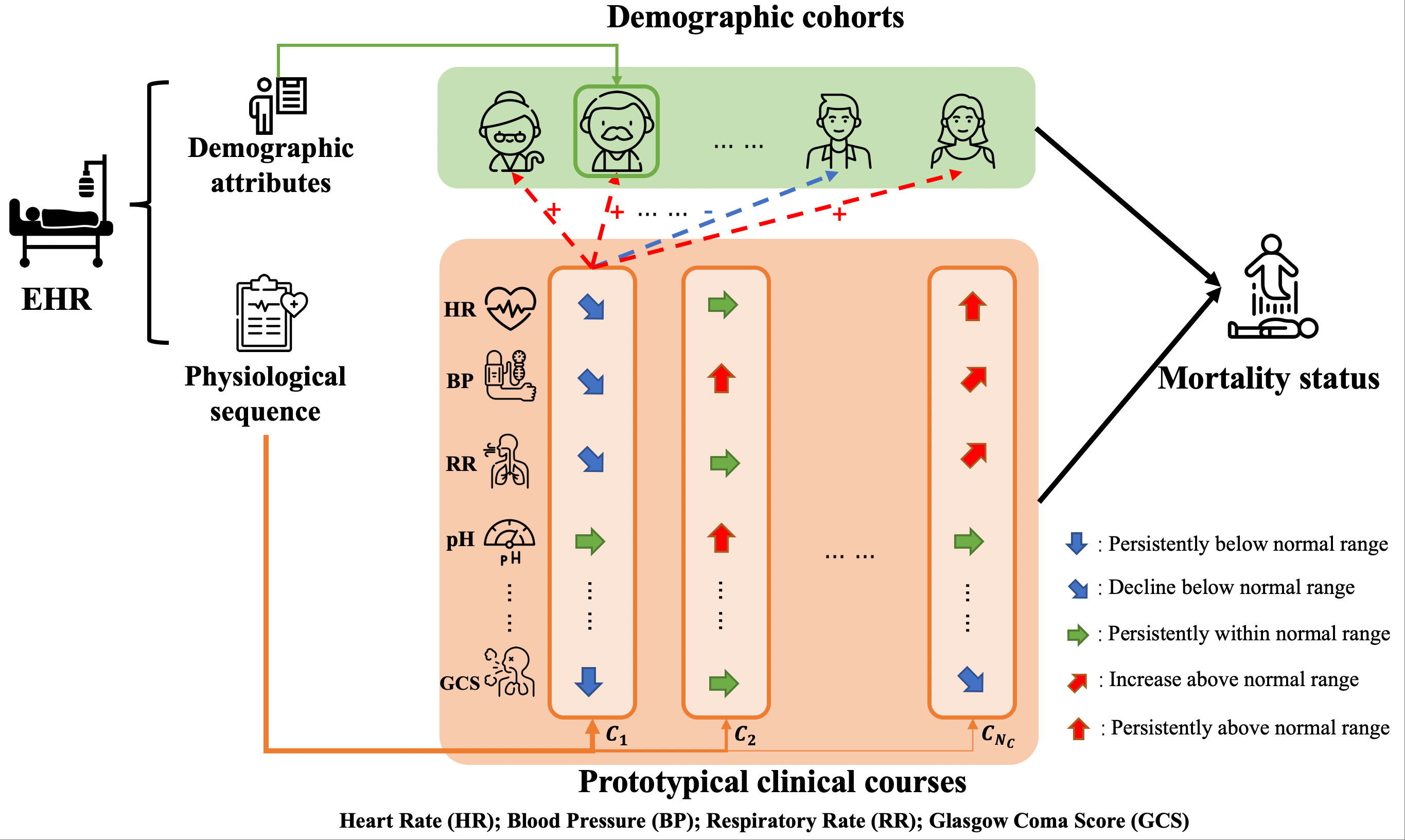}
    \caption{Illustration of the ICU decision-making practices.}
    \label{fig:inter_para}
\end{figure}

To maintain predictive performance of DL models without compromising interpretability in the high-stakes ICU environment, an ideal solution is to develop an intrinsically interpretable DL model whose reasoning logic closely aligns with the ICU decision-making practices. An illustration of the ICU decision-making practices is provided in Figure~\ref{fig:inter_para}, where the EHR for an ICU admission consists of the patient's demographic attributes and physiological sequence. First, faced with the complexity inherent in the multivariate physiological sequence, where multiple interdependent variables evolve over time, a doctor assesses the mortality risk by identifying the existing strengths of various \textbf{prototypical clinical courses} within the sequence~\citep{bhattacharyay2023mining}. Each prototypical clinical course represents a typical temporal pattern of physiological features and offers important insight into the patient's mortality risk~\citep{wang2020clinical}. For instance, the prototypical clinical course $C_1$ in Figure~\ref{fig:inter_para} is characterized by persistently low Glasgow Coma Scale (GCS) scores (a sign of deep coma) along with extended periods of severely reduced respiratory rates (evidence of respiratory failure). This clinical course is associated with high mortality risk as it exemplifies critical deterioration in neurological and pulmonary systems that strongly predict adverse ICU outcomes~\citep{fathi2022association}. A strong presence of the clinical course $C_1$ in the patient's physiological sequence in Figure~\ref{fig:inter_para} thus indicates the patient is facing a high mortality risk. This clinical course identification procedure offers a structured approach to analyzing and interpreting patients’ physiological data in relation to mortality risk, presenting a promising way of enhancing both the predictive performance and interpretability of the model. Also, the doctor estimates the implied mortality risk of each clinical course while accounting for \textbf{demographic heterogeneity}, which means the same clinical course may be associated with different risks across demographic cohorts. Specifically, the doctor first assigns the patient to a cohort based on the patient's demographic attributes and then assesses how the implied mortality risk of each clinical course should be adjusted according to the assigned cohort. This step is illustrated via the dashed red and blue arrows in Figure~\ref{fig:inter_para}, which indicate that the clinical course $C_1$ poses a higher mortality risk for older male patients compared to other cohorts~\citep{gonccalves2023critically}. By integrating the identified existing strengths of prototypical clinical courses with cohort-specific risk implications, the doctor forms a comprehensive and personalized assessment of the patient’s mortality status at discharge. The incorporation of demographic heterogeneity into the model's reasoning process enables it to account for cohort-specific risk adjustments, leading to more clinically meaningful interpretations and more accurate, personalized mortality predictions. Moreover, a key component of the ICU decision-making practices is \textbf{prognostication}, i.e., the forward-looking process of estimating a patient’s near-future health state from historical observations during admission~\citep{rencic2020situated}. Here, health state refers to the patient’s overall clinical condition at a given time, reflecting the combined severity of illness across multiple physiological systems. Unlike the previous two components, prognostication is future-oriented, focusing on anticipating how the health state will evolve rather than assessing or adjusting current risks. Since the mortality status at discharge, i.e., the prediction target in ICU mortality prediction, is closely related to the patient’s future health state, we embed this forward-looking perspective into model reasoning to enhance predictive performance. In conclusion, grounding the model's reasoning process in the ICU decision-making practices yields two key advantages. First, incorporating clinical insights from these practices into model design enables state-of-the-art predictive accuracy, consistent with established IS practices of integrating domain knowledge to enhance model performance ~\citep{abbasi2018text,he2019mobile,lee2024explainable,zhang2025ketch}. Second, the alignment of the model's intrinsic interpretability with the ICU decision-making practices increases transparency and trustworthiness. This alignment strengthens practitioner acceptance, facilitates clinical adoption, and ultimately amplifies real-world impact in critical care~\citep{bienefeld2023solving}.

Motivated by the above analysis, our research objective is to build an effective and intrinsically interpretable DL model for ICU mortality prediction, whose reasoning process should incorporate three key elements of the ICU decision-making practices: clinical course identification, demographic heterogeneity, and prognostication awareness.
By capturing these crucial aspects, state-of-the-art predictive performance can be achieved along with clinically meaningful interpretations. However, these three essential elements are seldom considered simultaneously and integrated with intrinsic interpretability in existing interpretable healthcare prediction methods, regardless of whether they are post hoc or intrinsic.
On the one hand, post hoc methods, while partially addressing demographic heterogeneity, only explain black-box DL models through local approximations. In this case, they do not reveal the actual reasoning logic of the model as intrinsic interpretations do, which may limit clinical trust and compromise regulatory compliance~\citep{hyland2020early,kim2023rolex}.
On the other hand, attention-based models provide intrinsic interpretations and account for demographic heterogeneity via learned attention weights~\citep{ma2020concare,ma2023mortality}. 
Yet, neither approach interprets model reasoning in terms of prototypical clinical courses, a cornerstone of the ICU decision-making practices.

One conceptually appealing approach to our objective is Prototype Learning, which offers intrinsic interpretability by interpreting predictions with reference to representative prototypes learned from training data, each reflecting a class-specific typical pattern. In our context of ICU mortality prediction, the key concepts of prototypical clinical courses and demographic cohorts can be recognized as distinct sets of prototypes to support interpretation.
Nevertheless, incorporating critical elements of the ICU decision-making practices into prototype learning introduces significant challenges. First, existing approaches generally operate on a single data modality, such as images~\citep{chen2019looks}, texts~\citep{hong2023protorynet}, or univariate time series~\citep{ming2019interpretable}.  In contrast, our task requires predicting and interpreting ICU mortality status based on both multivariate physiological sequences and static demographic attributes within EHRs. Yet, prototype learning models tailored to handle this multimodal data structure remain underexplored. Second, conventional prototype learning methods typically capture only the influence of a set of unimodal prototypes. When accounting for demographic heterogeneity, however, it becomes essential to model interactions between clinical course prototypes and demographic cohort prototypes, which originate from different modalities. This need reflects a universal principle in clinical practice: the same medical condition may have different implications depending on patient demographics~\citep{gabler2009dealing,varadhan2013framework}. However, the design of principled mechanisms to capture and integrate such cross-modal prototype interactions remains an open problem. Finally, the forward-looking nature of prognostication indicates that historical health states should provide predictive cues about future states. According to the clinical course identification process adopted by ICU doctors to initiate assessments, the health state at each period is represented as the existing strengths of different prototypical clinical courses reflected in the physiological data. From this perspective, the prognostication process suggests that these clinical-course-based representations should be predictable from preceding observations. Nevertheless, existing prototype learning frameworks offer no explicit means of operationalizing such temporal predictability.

To address the challenges, we propose \textbf{ProtoDoctor}, a novel prototype learning model for ICU mortality prediction. Unlike existing interpretable healthcare prediction models and current prototype learning methods, ProtoDoctor provides intrinsic interpretations while integrating all three essential elements of the ICU decision-making practices: clinical course identification, demographic heterogeneity, and prognostication awareness. To achieve this, ProtoDoctor builds on the prototype learning paradigm and introduces two methodological contributions: the Prognostic Clinical Course Identification (PCCI) module and the Demographic Heterogeneity Recognition (DHR) module. Specifically, the PCCI module learns clinical course prototypes for clinical course identification and incorporates a novel regularization strategy to capture prognostication awareness. The DHR module accounts for demographic heterogeneity by introducing demographic cohort prototypes and a novel interaction mechanism. Extensive empirical evaluations demonstrate that ProtoDoctor outperforms both existing interpretable healthcare prediction models and state-of-the-art prototype learning approaches adapted to this task. Human evaluation further validates that the interpretations generated by ProtoDoctor are more clinically meaningful and helpful for ICU practice compared to established interpretable baselines.

\section{Literature Review}

Two streams of research are closely related to our study: existing methods for interpretable healthcare prediction and state-of-the-art prototype learning models that are adaptable for interpretable ICU mortality prediction. 
In this section, we first review each stream of related studies and then summarize the key novelties of our work.

\subsection{HIT and ICU Mortality Prediction}\label{subsec:inter_HP}
Our study falls within the health information technology (HIT) domain and the related area of healthcare predictive analytics (HPA) in IS research~\citep{baird2020mis}. HIT encompasses \textit{``an array of technologies to store, share, and analyze health information''}. 
The proliferation of EHRs has yielded vast amounts of clinical data, creating opportunities for IS scholars to develop advanced IT artifacts to address diverse HPA challenges~\citep{morid2023time} such as adverse event prediction \citep{lin2021first}, length-of-stay prediction \citep{wang2022high,guo2024explainable}, and readmission prediction \citep{bardhan2015predictive, xie2021readmission}. In this study, we focus on the critical task of ICU mortality prediction and contribute to the HIT domain with a novel and interpretable solution for this significant task. Early approaches to ICU mortality prediction utilized prognostic scoring systems to analyze patients' clinical features in EHRs. Examples included widely used scores like the SAPS~\citep{le1984simplified}, the APACHE~\citep{zimmerman2006acute}, and the SOFA~\citep{vincent1996sofa}.
Although these scoring systems deliver transparent, interpretable frameworks for ICU practitioners, they are fundamentally calibrated for population‑level outcome estimation rather than individualized prediction. Their reliance on static measurements and omission of dynamic, context-specific variables prevents them from capturing the distinct temporal patterns present in each patient’s EHR. This limitation directly undermines their predictive precision at the individual level, thereby constraining their applicability to patient-specific decision-making in critical care. To address these limitations, recent studies have leveraged deep learning (DL) models to capture nuanced, patient-specific temporal patterns in longitudinal EHR data, resulting in substantial improvements in prediction accuracy~\citep{morid2023time, harutyunyan2019multitask,bednarski2022temporal,yu2024smart}. These studies formulate ICU mortality prediction as a binary classification problem and train a classifier from historical ICU admission records, each comprising EHR data and the corresponding mortality status at discharge. The trained classifier can then be adopted to predict the mortality status at discharge for new ICU admissions. For instance, \cite{harutyunyan2019multitask} compared multiple Recurrent Neural Network (RNN)-based approaches to capture temporal patterns in EHRs for ICU mortality prediction and concluded that the multitask channel-wise LSTM method achieved the best performance.
\citet{bednarski2022temporal} proposed a novel patient representation method that captures multi-scale temporal patterns within EHRs using temporal convolutional networks with different kernel sizes. SMART~\citep{yu2024smart} addressed the challenge of missing EHR data using an attention mechanism.
Collectively, these DL-based approaches have consistently outperformed traditional scoring systems in ICU mortality prediction.

While these DL-based approaches have achieved substantial gains in predictive accuracy, their limited interpretability poses a major barrier to adoption in high-stakes ICU settings.
Accordingly, research on interpretable DL models for healthcare prediction has emerged along two main directions: post-hoc interpretable methods and intrinsically interpretable approaches. Post-hoc methods explain a black-box predictor by locally approximating its input–output behavior with a simpler, more interpretable surrogate model~\citep{caicedo2019iseeu, hyland2020early, thorsen2020dynamic, qiu2022interpretable, kim2023rolex}. Popular post-hoc techniques, such as DeepLIFT~\citep{caicedo2019iseeu} and SHAP~\citep{hyland2020early, thorsen2020dynamic, qiu2022interpretable}, generated time-sensitive importance scores for individual clinical features, which indicate how strongly each feature at a given time point influences the predicted mortality risk and, in some cases, specify whether the influence increases or decreases the risk.
More recently, \citet{kim2023rolex} proposed ROLEX, a post-hoc framework that provides robust patient-level explanations in formats like feature importance scores and local decision rules.
While post-hoc explanations can be helpful, they may not faithfully reflect the actual reasoning process of the underlying model. This concern highlights the need for intrinsically interpretable approaches that make their reasoning logic transparent by design~\citep{rudin2019stop}.

In response, a growing body of work has focused on intrinsically interpretable methods, which aim to reveal the model's reasoning logic through specifically designed model components. Many recent approaches achieve this by incorporating attention mechanisms, which highlight important input features by assigning them higher attention weights~\citep{choi2016retain, ma2017dipole, ma2020concare, ma2023mortality, guo2024explainable}. Accordingly, features with higher attention weights are interpreted as having a greater influence on the predicted outcome.
For example, RETAIN~\citep{choi2016retain} employed a two-level attention architecture to determine which past hospital visits and which clinical variables within those visits are most relevant for predicting outcomes. Similarly, Dipole~\citep{ma2017dipole} applied an attention-based bidirectional RNN to quantify the importance of prior visits for diagnosis prediction.
Moreover, when incorporating demographic data within EHRs, recent studies, such as ConCare~\citep{ma2020concare}, AICare~\citep{ma2023mortality}, and the framework proposed by~\citet{guo2024explainable}, extended attention mechanisms to model interactions between static demographics and dynamic physiological data.
For instance, ConCare quantified the contributions of feature interactions by modeling their interdependencies through a self-attention–based RNN. These approaches not only improve predictive accuracy but also enhance interpretability by revealing how features jointly influence outcomes.

Despite these advantages, attention-based methods face several notable limitations. First, \citet{serrano2019attention} demonstrated that attention weights do not necessarily indicate a model’s true reasoning logic. In other words, higher attention weights assigned to input features do not guarantee stronger correlations with the predicted class, which limits the reliability of attention as an interpretation mechanism. Moreover, prior IS research has emphasized that model interpretations should mirror the way domain experts communicate insights~\citep{kim2006effects,abbasi2018text,lee2024explainable}. As outlined in the Introduction, effective ICU mortality prediction requires embedding three essential elements (i.e., clinical course identification, demographic heterogeneity, and prognostication awareness) into the model’s reasoning process. While attention mechanisms can partially address demographic heterogeneity, they fall short of simultaneously integrating all three elements. This gap has fueled growing interest in prototype learning, an emerging class of intrinsically interpretable methods with the potential to incorporate all these elements into model reasoning.

\subsection{Prototype Learning}
Prototype learning offers a promising path towards building intrinsically interpretable models that can embed all three critical elements of the ICU decision-making practices directly into a model’s reasoning. A pioneering study introduced ProtoPNet, a novel prototype-based framework for interpretable image recognition~\citep{chen2019looks}. ProtoPNet incorporates a prototype layer after the convolutional layers, where each prototype encodes a representative part of a specific class. By matching image embeddings at different locations to these prototypes, the model provides transparent reasoning for its predictions. For example, users can understand that a bird is classified as a ``clay-colored sparrow" because its head shape and wing pattern closely resemble those of the learned prototypes for that species. Since the introduction of ProtoPNet, prototype learning has also been extended to support intrinsic interpretability in various scenarios such as video classification \citep{trinh2021interpretable}, graph classification \citep{zhang2022protgnn, seo2024interpretable}, federated learning \citep{tan2022fedproto}, and reinforcement learning \citep{kenny2023towards, ragodos2022protox}. In parallel, various enhancements have been proposed to further improve interpretability. For example, \citet{nauta2021neural} combined decision trees with prototype learning to enable hierarchical reasoning. \citet{rymarczyk2022interpretable} introduced ProtoPShare, which allows some attributes to be shared across prototypes to reduce redundancy. \citet{bontempelli2023concept} proposed ProtoPDebug, an effective concept-level debugger that incorporated human feedback into the prototype generation process, offering a promising tool for trustworthy interactive learning.

In the context of ICU mortality prediction, where dynamic physiological sequences and static demographic attributes serve as inputs, prototype learning methods for interpretable sequential modeling are particularly relevant to our study \citep{ming2019interpretable, ni2021interpreting, kuang2023symptoms, xie2025care, hong2023protorynet, xie2024prototype}. For instance, \citet{ming2019interpretable} introduced ProSeNet, the first prototype-based interpretable framework for sequential data.  It makes predictions by comparing an input time-series with a set of class-associated prototypical sequences. Similarly, \citet{xie2024prototype} proposed PahNet, a prototype-based model for ECG classification, where each prototype represents a typical ECG sequence associated with a particular cardiac condition. In both cases, prototypes are interpreted as representative complete sequences observed in the training data. To enable more detailed interpretations, some studies define prototypes at a finer granularity. Specifically, SCNpro~\citep{ni2021interpreting} derived prototypes from local segments, enabling more detailed interpretations in time-series prediction tasks. ProtoryNet~\citep{hong2023protorynet} interpreted document classification through trajectories of sentence-level prototypes, linking each sentence to a representative prototype to improve interpretability. More recently, \citet{xie2025care} introduced a dual-level prototype framework that models temporal prototype progressions, thereby facilitating interpretable depression detection from sensor data. Building on these advances, prototype learning can be adapted in our setting to represent core concepts such as clinical courses and demographic cohorts, which provides a foundation for embedding the critical elements of the ICU decision-making practices into model reasoning.

\subsection{Key Novelties of Our Study}\label{subsec:innovation}
As discussed before, an ideal solution for interpretable ICU mortality prediction should achieve intrinsic interpretability while embedding three critical elements into its reasoning process: clinical course identification, demographic heterogeneity, and prognostication awareness. Our literature review shows that existing interpretable prediction methods all fall short in some of these aspects, as listed in Table~\ref{tb:methodcomp}. Post-hoc methods, while capable of accounting for demographic heterogeneity, provide only local approximations of a DL model’s input–output behavior and therefore lack intrinsic interpretability. Attention-based models can offer intrinsic interpretations and capture demographic heterogeneity through learned attention weights. However, neither post-hoc nor attention-based approaches can analyze and interpret the mortality risks implied by patients’ physiological data in a manner that mimics the clinical course identification process adopted by ICU doctors. Prototype-based methods are conceptually appealing for achieving intrinsic interpretability while potentially incorporating all three elements into model reasoning. However, several challenges remain unresolved.
First, although prototypes can be used to represent key concepts, e.g., clinical courses, to embed clinical course identification within model reasoning, it remains unclear how to apply prototype learning to multimodal inputs in our context of ICU mortality prediction.
Second, it remains an open problem to capture and integrate cross-modal prototype interactions to effectively address demographic heterogeneity. Moreover, none of the above methods incorporates prognostication awareness into model reasoning. To bridge these gaps, we propose ProtoDoctor, a novel prototype-based framework for ICU mortality prediction. It achieves intrinsic interpretability while integrating all three essential elements into model reasoning, thus achieving state-of-the-art predictive accuracy and delivering clinically meaningful interpretations.

\renewcommand{\arraystretch}{1.5} 
\begin{table}[t]
    \centering
    \fontsize{8}{10}\selectfont
    \caption{Comparison between Our Method and Existing Interpretable Methods.}
    \begin{tabular}{ll|>{\centering\arraybackslash}p{2cm}>{\centering\arraybackslash}p{2cm}>{\centering\arraybackslash}p{2cm}>{\centering\arraybackslash}p{2cm}}
        \toprule
        \multirow{2.3}{*}{Category} & \multirow{2.3}{*}{Representative Methods} & \multirow{2.3}{2cm}{\centering Intrinsic Interpretability} & \multicolumn{3}{c}{ICU Reasoning Alignment} \\
        \cmidrule(lr){4-6}
        & & & \parbox[c]{2cm}{\centering \fontsize{7}{7}\selectfont Clinical Course Identification} & 
                \parbox[c]{2cm}{\centering \fontsize{7}{7}\selectfont Demographic Heterogeneity} & 
                \parbox[c]{2cm}{\centering \fontsize{7}{7}\selectfont Prognostication Awareness} \\
        \midrule
        \multirow{2}{*}{Post-hoc}& SHAP~\citep{hyland2020early} &  &  & \checkmark &  \\
        & ROLEX~\citep{kim2023rolex} &  &  & \checkmark &  \\
        \midrule
        \multirow{2}{*}{Attention-based} & Concare~\citep{ma2020concare} & \checkmark &  & \checkmark &  \\
        & AIcare~\citep{ma2023mortality} & \checkmark &  & \checkmark & \\
        \midrule
        \multirow{2}{*}{Prototype-based} & ProtoryNet~\citep{hong2023protorynet} & \checkmark & \checkmark &  &  \\
        & PahNet~\citep{xie2024prototype} & \checkmark & \checkmark &  &  \\
        \midrule
        \multicolumn{2}{c|}{ProtoDoctor (Ours)} & \checkmark & \checkmark & \checkmark & \checkmark \\
        \bottomrule
    \end{tabular}
    \label{tb:methodcomp}
\end{table}

\section{Problem Formulation}
In this section, we formally define the interpretable ICU mortality prediction problem. Let $U$ denote the set of ICU admissions. The clinical record of each ICU admission $u$ consists of the patient's EHR and the corresponding in-hospital mortality status during the admission, denoted as $X^{(u)}$ and $y^{(u)}$, respectively.
Since a patient may experience multiple ICU admissions, this study focuses on predicting the mortality status for each individual admission.
Specifically, the mortality status at discharge $y^{(u)}$ is a binary variable, where $y^{(u)} = 1$ indicates death and $y^{(u)} = 0$ indicates survival at discharge.
The EHR $X^{(u)}$ encompasses the patient's demographic attributes $\mathcal{D}^{(u)}$ and physiological sequence $\mathcal{P}^{(u)}$, i.e., $X^{(u)}=<\mathcal{D}^{(u)},\mathcal{P}^{(u)}>$. The details of these components are described below.

\textbf{Demographic attributes.} A patient's demographic attributes capture essential demographic information that is often highly relevant to the patient's final mortality status during an ICU admission. One established attribute is age, as older patients typically face higher mortality risks. 
Given an admission $u$, the demographic attributes is denoted as $\mathcal{D}^{(u)}=\{\mathcal{D}^{(u)}_1, \dots, \mathcal{D}^{(u)}_{n_{\mathcal{D}}} \}$, where $\mathcal{D}^{(u)}_k$ is the $k$-th demographic attribute and $n_{\mathcal{D}}$ is the number of demographic attributes.

\textbf{Physiological sequence.} The physiological sequence contains continuously monitored data of critical physiological variables such as heart rate and glucose levels during the initial stage of an ICU admission. 
More specifically, for each admission $u$, the physiological sequence is recorded as a time series of physiological features over the first $T$ hours of the admission, denoted as $\mathcal{P}^{(u)}=[\mathcal{P}_1^{(u)},\dots,\mathcal{P}_T^{(u)}]$, where $\mathcal{P}_t^{(u)}= [\mathcal{P}_{t,1}^{(u)},\dots,\mathcal{P}_{t,n_\mathcal{P}}^{(u)}]$ collects the readings of $n_{\mathcal{P}}$ physiological variables at hour $t$.

Accordingly, we denote the overall dataset as $\mathbf{D} = \{(X^{(u)}, y^{(u)}) \mid u = 1, 2, \dots, |U|\}$, where $|U|$ is the total number of admissions. We now formally define the interpretable ICU mortality prediction problem as follows.

\begin{definition}[\textbf{Interpretable ICU Mortality Prediction}]
\label{def:prob}
Given a dataset $\mathbf{D}$ containing clinical records of historical ICU admissions, our objective is to develop a model that can accurately predict the mortality status at discharge $\hat{y}^{(v)}$ for a new admission $v$ based on the corresponding EHR $X^{(v)}$, while providing clinically meaningful interpretations as discussed in the Introduction.
\end{definition}

\section{Method}
\begin{figure}[h]
    \centering
    \includegraphics[width=1\linewidth]{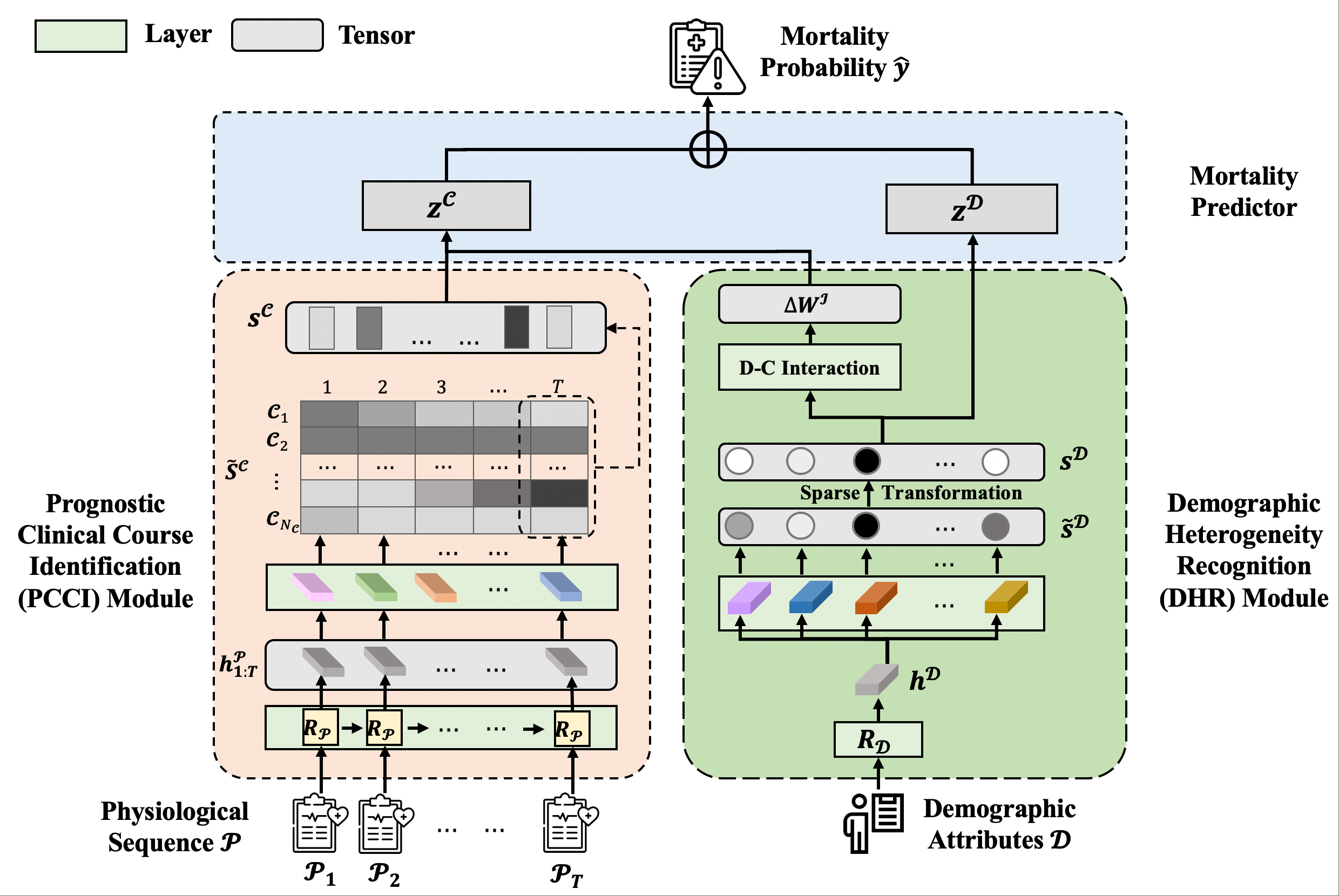}
    \caption{ProtoDoctor Architecture}
    \label{fig:framework}
\end{figure}

In this section, we propose \textbf{ProtoDoctor} (\underline{Proto}typical \underline{D}em\underline{o}graphic-and-Clinical-\underline{C}ourse-aware Ne\underline{t}w\underline{or}k), a novel framework for interpretable ICU mortality prediction. Figure~\ref{fig:framework} shows the overall architecture of our method, which is built upon three major components: the Prognostic Clinical Course Identification (PCCI) module, the Demographic Heterogeneity Recognition (DHR) module, and the Mortality Predictor. Given a patient's EHR, the PCCI module identifies the presence of prototypical clinical courses by first learning a set of clinical course prototypes and then quantifying their existing strengths within the patient's physiological sequence. To enhance model performance, this module incorporates a novel regularization term that emulates the ICU doctors' prognostication process. The DHR module accounts for demographic heterogeneity by adjusting the mortality risks associated with different clinical courses according to the patient’s attributed demographic cohort. It learns demographic cohort prototypes, assigns each patient to a specific cohort, and adjusts mortality risks accordingly. This novel cohort-based adjustment mechanism thus enables more personalized predictions and clinically meaningful interpretations. Finally, the Mortality Predictor predicts the patient's mortality status at discharge by integrating information on the attributed demographic cohort, the existing strengths of prototypical clinical courses, and the adjusted mortality risks associated with those clinical courses. The PCCI module and the DHR module constitute the major methodological novelties of our method. In the following discussion, we drop the symbol $u$ from most notations by focusing on the prediction of mortality status at discharge for one particular ICU admission. For convenience, we summarize key notations in Table~\ref{Table_Notation}. 

\begin{table}[h]
\centering
\caption{Notation}\label{Table_Notation}
\begin{tabular}{ll}
\toprule
Notation & Description \\
\hline
$\mathcal{P}$ & The physiological sequence recorded for one ICU admission, where $\mathcal{P}_t$ is \\
&the physiological features at hour $t$. \\
$\mathcal{D}$ & The patient's demographic attributes associated with one ICU admission. \\
$h^{\mathcal{P}}_t$ & The embedding vector of physiological sequence $\mathcal{P}$ up to hour $t$ . \\
$h^\mathcal{D}$ & The embedding vector of demographic attributes $\mathcal{D}$. \\
$p^{\mathcal{C}}_k$ & The prototype vector representing clinical course $k$. \\
$p^{\mathcal{D}}_m$ & The prototype vector representing demographic cohort  $m$. \\
$\tilde{\mathrm{s}}^{\mathcal{C}}_{t}$ & The health state vector at hour $t$, where $\tilde{s}^{\mathcal{C}}_{t,k}$ is the existing \\
 & strength of clinical course $k$ up to hour $t$, defined by Equation \ref{eq:ts}. \\
$\hat{\mathrm{s}}^{\mathcal{C}}_{t}$ & The predicted health state vector at hour $t$, defined by Equation \ref{eq:ep}. \\
$\tilde{\mathrm{s}}^{\mathcal{D}}$ & The demographic similarity vector, where $\tilde{s}^{\mathcal{D}}_m$ denotes the patient's matching\\
 & degree towards demographic cohort $m$, defined by Equation \ref{eq:sim_d} \\
$\mathrm{s}^{\mathcal{D}}$ & The sparse demographic attribution vector, defined by Equation \ref{eq:sparse}. \\

\bottomrule
\end{tabular}
\end{table}

\subsection{Prognostic Clinical Course Identification Module}\label{subsec:EPM}
Given a patient’s multivariate physiological sequence, it is challenging to analyze and interpret it for ICU mortality prediction in a manner consistent with the ICU decision-making practices. We address this challenge by drawing on the well-established concept of the clinical course.
\begin{definition}[\textbf{Clinical course}]
\label{def:course}
    The clinical course is an abstract concept describing the time-dependent progression of a patient’s medical condition. In the context of ICU mortality prediction, it represents a characteristic temporal pattern formed by the joint evolution of multiple physiological variables, offering critical insights into the patient’s mortality risk~\citep{marino1998icu}. 
\end{definition}

This definition underscores the importance of modeling temporal patterns across the entire physiological sequence rather than analyzing isolated observations independently. In practice, ICU doctors initiate mortality risk assessment by identifying prototypical clinical courses within the sequence~\citep{wang2020clinical,bhattacharyay2023mining}. To embed this procedure into model reasoning, we represent the patient’s health state at each period as a vector indicating the existing strengths of different prototypical clinical courses in the physiological sequence. We refer to this as the health state vector.
Additionally, a critical step in the ICU decision-making practices is prognostication, where doctors infer patients' near-future health states based on historical observations. Inspired by the forward-looking nature of this step, we aim to incorporate prognostication awareness into model reasoning. By doing so, we expect the health state vector to better reflect near-future health states, thus facilitating ICU mortality prediction.

\begin{figure}
    \centering
    \includegraphics[width=0.75\linewidth]{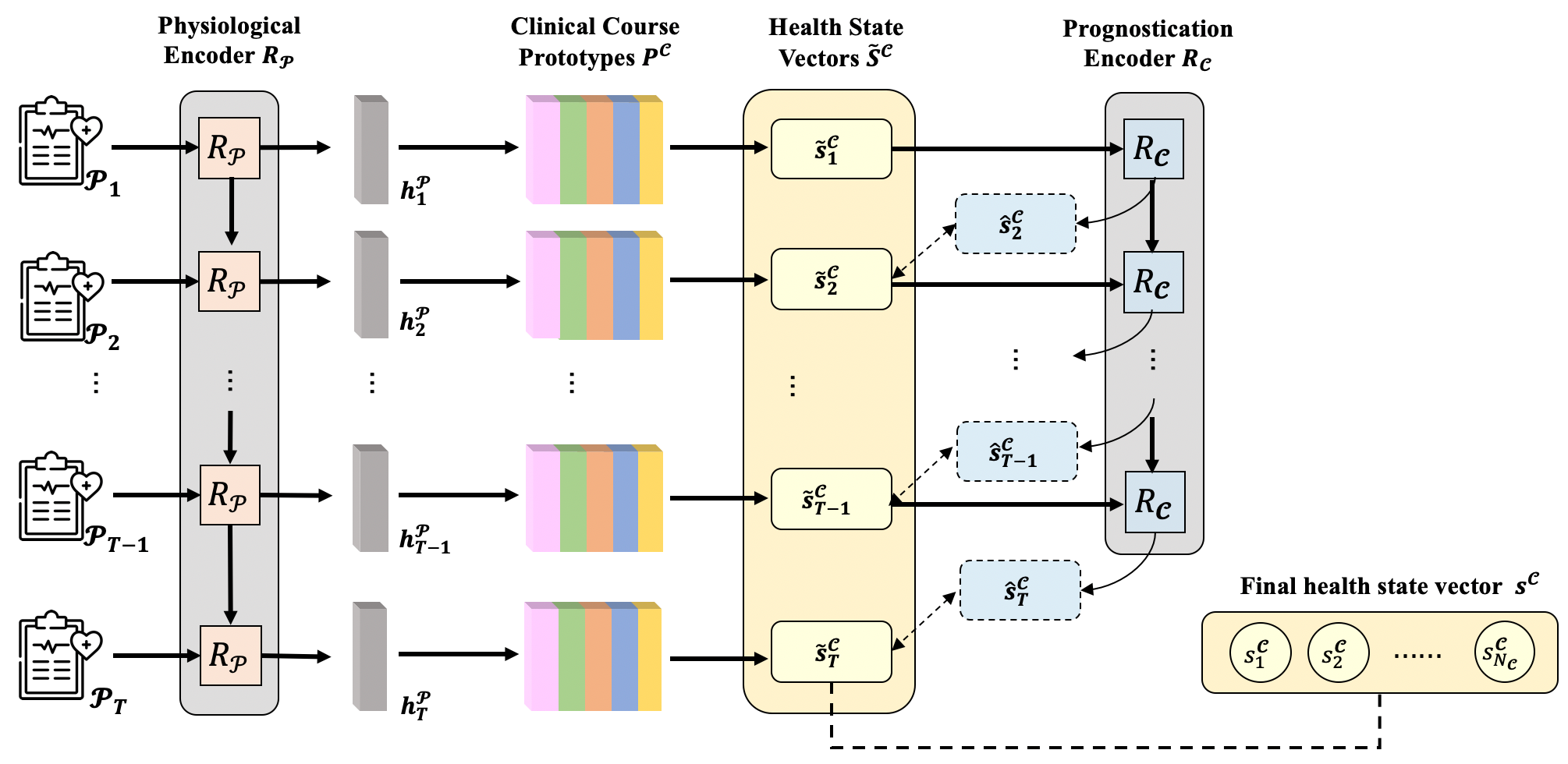}
    \caption{Architecture of the Prognostic Clinical Course Identification Module}
    \label{fig:EPM}
\end{figure}

To implement these ideas, the PCCI module first learns a set of clinical course prototypes, where each prototype represents a prototypical clinical course associated with either death or survival at discharge. The patient's health state at each period can then be represented as a vector of existing strengths of these prototypes to facilitate both prediction and interpretation. 
Finally, we introduce a novel regularization mechanism to incorporate prognostication awareness within model reasoning, which further enhances predictive performance on ICU mortality prediction.
The overall architecture of the PCCI module is shown in Figure~\ref{fig:EPM}.

\subsubsection{Clinical Course Identification.}
Given the physiological sequence $\mathcal{P} =[\mathcal{P}_1, \dots, \mathcal{P}_T]$ at the first $T$ hours of an ICU admission, we adopt a physiological encoder $R_\mathcal{P}$ to generate hourly embeddings $H^\mathcal{P}=[h^\mathcal{P}_1, \dots, h^\mathcal{P}_T]^T \in R^{T\times n_{\mathcal{P}}}$, where $h^{\mathcal{P}}_t \in R^{n_{\mathcal{P}}}$ is the embedding vector of physiological sequence $\mathcal{P}$ up to hour $t$ and $n_{\mathcal{P}}$ is the embedding dimension size. We define $H^\mathcal{P}$ as:
\begin{equation}\label{eq:enc_rp}
\begin{split}
    H^\mathcal{P} &= [h^\mathcal{P}_1, \dots, h^\mathcal{P}_T]\\
    & = R_\mathcal{P}([\mathcal{P}_1,\dots,\mathcal{P}_T]).
\end{split}
\end{equation}
The physiological encoder $R_\mathcal{P}$ is an RNN layer based on the architecture from~\citet{harutyunyan2019multitask}, with specifications detailed in the Benchmark Methods section.

To identify what typical clinical courses are exhibited within the patient's physiological sequence, we first learn a set of $N_{\mathcal{C}}$ clinical course prototypes, each representing a prototypical clinical course associated with death or survival at discharge.
These prototypes, denoted as $P^{\mathcal{C}}=\{p^{\mathcal{C}}_1, \dots, p^{\mathcal{C}}_{N_\mathcal{C}} \}$, are learned as trainable parameters and embedded in the same latent space as the physiological embedding vector $h^\mathcal{P}_t$.
Each prototype vector $p^{\mathcal{C}}_k$ for $k=1,\dots, N_\mathcal{C}$ is interpreted and visualized by the closest physiological sequence in the embedding space, as described in \ref{subsec:PI}. Then, we define $\tilde{s}^{\mathcal{C}}_{t,k}$ to represent the existing strength of clinical course $k$ up to hour $t$. This value is computed as the similarity between the patient's physiological embedding vector, $h_t^{\mathcal{P}}$, and the corresponding prototype vector $p^{\mathcal{C}}_k$:
\begin{equation}\label{eq:ts}
    \tilde{s}^{\mathcal{C}}_{t,k} = \sigma\big(\varphi \cdot \cos(h^{\mathcal{P}}_t, p^{\mathcal{C}}_k)\big),
\end{equation}
where $\cos(x,y)=x^T y / (||x|| \cdot ||y||)$ is the cosine similarity between the input vectors $x$ and $y$, $\sigma(z)=1/\big(1+\exp(-z)\big)$ is the sigmoid function, and $\varphi$ is a positive scaling factor (e.g., $\varphi=5$). The definition of $\tilde{s}^{\mathcal{C}}_{t,k}$ ensures that its value falls within $(0,1)$ with a higher value indicating a stronger existence of clinical course $k$ up to hour $t$. The health state vector $\tilde{\mathrm{s}}^{\mathcal{C}}_t$ at hour $t$ is then defined as the collection of similarity scores with respect to all clinical courses. Namely,
\begin{equation}\label{eq:physio_vec}
     \tilde{\mathrm{s}}^{\mathcal{C}}_t =[\tilde{s}^{\mathcal{C}}_{t,1},\dots,\tilde{s}^{\mathcal{C}}_{t,N_{\mathcal{C}}}].
\end{equation}
By aggregate these vectors $[\tilde{\mathrm{s}}^{\mathcal{C}}_1,\dots,\tilde{\mathrm{s}}^{\mathcal{C}}_T]$ over time, we can capture the temporal progression of the patient's health state from the perspective of clinical courses.

\subsubsection{Prognostication-aware Regularization.}
To further improve model performance, we propose a novel regularization mechanism inspired by the prognostication step in the ICU decision-making practices:
\begin{definition}[\textbf{Prognostication}]
\label{def:prognostication}
    \textit{Prognostication} is a forward-looking step in the ICU decision-making practices, where doctors analyze a patient’s historical records to estimate the near-future health state. Situated cognition theory~\citep{rencic2020situated} further requires that doctors should \textit{``continually collecting and interpreting patient data, prognosticating, ..., to improve a patient's well-being"} given the emerging clinical data in the ICU environment.
\end{definition}

The definition indicates that a patient’s near-future health state can be inferred from historical observations. Also, this forward-looking process should be continuously updated in response to emerging clinical data. We incorporate prognostication awareness into model reasoning by enforcing that the patient’s health state at each period is predictable from preceding states. This constraint guides the model to capture temporal dependencies that are indicative of future conditions, thereby improving the accuracy of ICU mortality prediction.

To operationalize this predictability constraint, we introduce an RNN-based prognostic encoder $R_{\mathcal{C}}$ to predict the health state vector $\tilde{\mathrm{s}}^{\mathcal{C}}_t$ based on all preceding health state vectors $\tilde{\mathrm{s}}^{\mathcal{C}}_{1:t-1}$:
\begin{equation}\label{eq:ep}
    \hat{\mathrm{s}}^{\mathcal{C}}_{t} = R_{\mathcal{C}}(\tilde{\mathrm{s}}^{\mathcal{C}}_{1:t-1}), t=2,\dots,T.
\end{equation}
We then impose a regularization term that minimizes the discrepancy between the predicted and observed health state vectors:
\begin{equation}\label{eq:Lp}
    L_p^{\mathcal{C}} = \frac{1}{T-1} \sum_{t=2}^T ||\hat{s}_t^{\mathcal{C}}-\tilde{s}_t^{\mathcal{C}}||^2_2,
\end{equation}
where $\hat{s}_t^{\mathcal{C}}$ and $\tilde{s}_t^{\mathcal{C}}$ denote the predicted and observed health state vectors at hour $t$, respectively. Guided by the prognostication process, our method performs step-ahead prediction on the interpretable health state vector derived from clinical course prototypes. Unlike existing healthcare prediction studies that enhance prediction by introducing auxiliary tasks on raw features or discrete clinical events (e.g., decompensation prediction)~\citep{harutyunyan2019multitask,lee2019dynamic}, we focus on a clinically meaningful representation in the latent space. This representation is less sensitive to noise in raw features and better captures the subtle evolution of patient conditions, thereby providing a more robust and effective regularization mechanism for ICU mortality prediction.

\subsection{Demographic Heterogeneity Recognition Module}\label{subsec:GAM}
The DHR module is designed to capture demographic heterogeneity, a well-documented phenomenon in which the same clinical course can imply different mortality risks across patient cohorts~\citep{vallet2023impact}. To achieve this, the module introduces demographic cohort prototypes and generates an interpretable demographic attribution vector for each patient, mirroring how ICU doctors assign patients to specific cohorts. It then employs a novel interaction component to produce cohort-specific mortality risk adjustments, enabling the same clinical course to carry different implications depending on the patient’s cohort. These adjustments yield personalized mortality risks associated with different clinical courses for each patient.

\subsubsection{Demographic Cohort Attribution.}
Given the patient's demographic attributes $\mathcal{D}$, we first leverage a demographic encoder $R_{\mathcal{D}}$ to generate a corresponding embedding vector $h^{\mathcal{D}} \in R^{n_\mathcal{D}}$, where $n_\mathcal{D}$ is the embedding dimension size:
\begin{equation}\label{eq:enc_rd}
    h^{\mathcal{D}} = R_{\mathcal{D}}(\mathcal{D}).
\end{equation}
$R_{\mathcal{D}}$ is implemented as an MLP layer, with details provided in the Benchmark Methods section. To attribute the patient to a specific demographic cohort, we first learn a set of $N_{\mathcal{D}}$ demographic cohort prototypes, denoted as $P^{\mathcal{D}}=\{p^{\mathcal{D}}_1, \dots, p^{\mathcal{D}}_{N_\mathcal{D}} \}$. Each prototype vector $p^{\mathcal{D}}_m \in R^{n_{\mathcal{D}}} (m = 1, \dots, N_{\mathcal{D}})$ is a trainable vector representing a typical demographic cohort. To quantify how closely a patient matches demographic cohort $m$, we compute the similarity $\tilde{s}^{\mathcal{D}}_m$ between the patient's demographic embedding vector $h^{\mathcal{D}}$ and the prototype vector $p^{\mathcal{D}}_m$:
\begin{equation}\label{eq:sim_d}
    \tilde{s}^{\mathcal{D}}_m = \sigma\big({\varphi \cdot \cos(h^{\mathcal{D}}, p^{\mathcal{D}}_m)}\big),
\end{equation}
where $\tilde{s}^{\mathcal{D}}_m \in (0,1)$, and higher values indicate that the patient exhibits a stronger match towards demographic cohort $m$. The similarity scores are then assembled into the demographic similarity vector:
\begin{equation}\label{eq:demo_sim}
\tilde{\mathrm{s}}^{\mathcal{D}}=[\tilde{s}^{\mathcal{D}}_1,\dots,\tilde{s}^{\mathcal{D}}_{N_{\mathcal{D}}}].
\end{equation}
To attribute the patient to a single demographic cohort, we apply the sparse transformation from ProtoryNet \citep{hong2023protorynet} to the similarity vector, producing the sparse demographic attribution vector $\mathrm{s}^{\mathcal{D}}$:
\begin{equation}\label{eq:sparse}
    \mathrm{s}^{\mathcal{D}} = Softmax(\tilde{s}^{\mathcal{D}}_1/\tau, \dots, \tilde{s}^{\mathcal{D}}_{N_\mathcal{D}}/\tau),
\end{equation}
with $\tau$ being a small constant (e.g., $10^{-6}$) that ensures $\mathrm{s}^{\mathcal{D}}$ approximates a one-hot vector. The prototype corresponding to the highest similarity score in $\tilde{s}^{\mathcal{D}}$ is then selected as the activated demographic cohort prototype, representing the patient’s assigned demographic cohort.

\subsubsection{D-C Interaction.}
To account for the heterogeneous mortality risks associated with a single clinical course across different demographic cohorts, we introduce a novel \textit{D-C Interaction} component, which explicitly integrates demographic heterogeneity into ProtoDoctor's reasoning process.
Specifically, we define a learnable interaction contribution matrix $\Delta \mathbb{W}^{\mathcal{I}} \in  R^{N_{\mathcal{C}} \times N_{\mathcal{D}}}$, where $N_{\mathcal{C}}$ and $N_{\mathcal{D}}$ denote the number of clinical course prototypes and demographic cohort prototypes, respectively. Each element $\Delta w^{\mathcal{I}}_{k,m}$ represents the mortality risk adjustment for clinical course $k$ when the patient belongs to demographic cohort $m$:

\begin{equation}
    {\Delta\mathbb{W}^{\mathcal{I}}} = 
\left[
\begin{array}{cccc}
\Delta w^{\mathcal{I}}_{1,1} & \Delta w^{\mathcal{I}}_{1,2} & \cdots & \Delta w^{\mathcal{I}}_{1,N_{\mathcal{D}}} \\
\Delta w^{\mathcal{I}}_{2,1} & \Delta w^{\mathcal{I}}_{2,2} & \cdots & \Delta w^{\mathcal{I}}_{2,N_{\mathcal{D}}} \\
\vdots & \vdots & \ddots & \vdots \\
\Delta w^{\mathcal{I}}_{N_{\mathcal{C},1}} & \Delta w^{\mathcal{I}}_{N_{\mathcal{C},2}} & \cdots & \Delta w^{\mathcal{I}}_{N_{\mathcal{C}},N_{\mathcal{D}}}
\end{array}
\right].
\end{equation}
Each row of the ${\Delta\mathbb{W}^{\mathcal{I}}}$ thus encodes how mortality risk adjustments for a specific clinical course vary across demographic cohorts.
Given a patient’s demographic attribution vector $\mathrm{s}^{\mathcal{D}}$, the personalized risk adjustment vector $\Delta W^{\mathcal{I}} \in \mathbb{R}^{N_{\mathcal{C}}}$ is computed as:
\begin{equation}\label{eq:adjust}
    \Delta W^{\mathcal{I}} =  \Delta\mathbb{W}^{\mathcal{I}} \mathrm{s}^{\mathcal{D}}.
\end{equation}
This vector specifies patient-specific mortality risk adjustments for each clinical course, conditioned on the attributed demographic cohort (if $\mathrm{s}^{\mathcal{D}}_m\approx 1$ then $\Delta W^{\mathcal{I}}$ is contributed almost entirely by the $m$th column of $\Delta\mathbb{W}^{\mathcal{I}}$).
These adjustments are subsequently combined with the base mortality risks to produce personalized ICU mortality risk predictions. Details of this integration are provided in Section~\ref{subsec:LO}.

\subsection{Mortality Predictor}\label{subsec:LO}
The Mortality Predictor estimates the patient’s mortality status at discharge by integrating three components: the health state vector at the final hour ($\tilde{\mathrm{s}}^{\mathcal{C}}_T$, denoted as $\mathrm{s}^{\mathcal{C}}$ for simplicity); the demographic attribution vector ($\mathrm{s}^{\mathcal{D}}$); and the risk adjustment vector ($\Delta W^{\mathcal{I}}$). These components jointly produce two complementary mortality risk scores, $z^{\mathcal{C}}$ and $z^{\mathcal{D}}$, as illustrated in Figure~\ref{fig:framework}. The first score $z^{\mathcal{C}}$ captures the patient's clinical-course-related mortality risks by combining the existing strengths of different clinical courses within the patient's physiological sequence with the personalized mortality risk associated with each clinical course. Formally:
\begin{equation}\label{eq:Zc}
\begin{split}
z^{\mathcal{C}} &= {\tilde W}^{\mathcal{C}} \cdot \mathrm{s}^{\mathcal{C}}\\
&= (\mathbb{W}^{\mathcal{C}}+\Delta W^{\mathcal{I}}) \cdot \mathrm{s}^{\mathcal{C}}\\
&=\sum_{k=1}^{N_{\mathcal{C}}} (\mathbb{W}^{\mathcal{C}}_k+\Delta W^{\mathcal{I}}_k) \cdot \mathrm{s}^{\mathcal{C}}_k.
\end{split}
\end{equation}
where $\mathbb{W}^{\mathcal{C}} \in R^{N_{\mathcal{C}}}$ is a learnable weight vector with each element $\mathbb{W}^{\mathcal{C}}_k$ representing the base mortality risk implied by clinical course $k$, while $\mathrm{s}^{\mathcal{C}}$ reflects the existing strengths of different clinical courses at the final hour $T$. The additive term $\Delta W^{\mathcal{I}}$ incorporates cohort-specific adjustments, enabling personalized clinical-course-based risk estimation. Notably, this score is computed as the sum of risk contributions from all $N_{\mathcal{C}}$ clinical courses. Each term $(\mathbb{W}^{\mathcal{C}}_k+\Delta W^{\mathcal{I}}_k) \cdot \mathrm{s}^{\mathcal{C}}_k$ represents the adjusted mortality risk implied by clinical course $k$, weighted by its existing strength within the physiological sequence of the given patient. This decomposition clarifies how both the general population-level risks and patient-specific adjustments jointly determine the overall clinical-course-related mortality score. The second score, $z^{\mathcal{D}}$, quantifies the mortality risk from a demographic perspective:
\begin{equation}
\begin{split}
z^{\mathcal{D}} &= \mathbb{W}^{\mathcal{D}} \cdot \mathrm{s}^{\mathcal{D}} \\
&= \sum_{m=1}^{N_{\mathcal{D}}}\mathbb{W}^{\mathcal{D}}_m \cdot \mathrm{s}^{\mathcal{D}}_m.
\end{split}
\end{equation}
$\mathbb{W}^{\mathcal{D}} \in R^{N_{\mathcal{D}}}$ is a learnable vector where each element represents the mortality risk associated with a demographic cohort. The vector $\mathrm{s}^{\mathcal{D}}$ approximates a one-hot vector, with one element typically close to one and the rest near zero. Thus, when $\mathrm{s}^{\mathcal{D}}_m\approx 1$, the demographic risk score $z^{\mathcal{D}}$ is determined almost entirely by the corresponding risk weight $\mathbb{W}^{\mathcal{D}}_m$.
Finally, the predicted mortality status $\hat{y}$ is obtained by aggregating the clinical-course-based score $z^{\mathcal{C}}$ and the demographic-based score $z^{\mathcal{D}}$, followed by a sigmoid activation:

\begin{equation}\label{eq:Final}
\begin{split}
\hat{y} &= \sigma (z^{\mathcal{C}}+z^{\mathcal{D}}) \\ 
        &= \sigma ({\tilde W}^{\mathcal{C}} \cdot s^{\mathcal{C}}+ \mathbb{W}^{\mathcal{D}} \cdot \mathrm{s}^{\mathcal{D}}) \\ 
\end{split}
\end{equation}

\subsection{ProtoDoctor}
Given the clinical record $(X^{(u)}, y^{(u)}) = (\langle \mathcal{D}^{(u)}, \mathcal{P}^{(u)} \rangle, y^{(u)})$ of an admission $u \in U$, ProtoDoctor predicts the patient's mortality status at discharge according to the following interpretable reasoning process. The model first quantifies the existing strengths of different prototypical clinical courses within the patient’s physiological sequence at the final hour $T$ (Equation~\ref{eq:physio_vec}). Based on the patient’s demographic attributes, it then assigns the patient to a specific demographic cohort(Equation~\ref{eq:sparse}). 
Next, personalized risk adjustments are applied for different clinical courses (Equation~\ref{eq:adjust}). Finally, these components are integrated to estimate the mortality status at discharge (Equation~\ref{eq:Final}).

Next, we illustrate how to learn the model parameters based on the training dataset.
Let $\Phi$ denote the complete set of model parameters in ProtoDoctor, we have:
\begin{equation}
    \Phi = \{\Theta, \mathbb{P}, \mathbb{W} \}.
\end{equation}
Here, $\Theta=\{\theta^{\mathcal{P}}, \theta^{\mathcal{C}},\theta^{\mathcal{D}}\}$ comprises the parameters of the physiological encoder $R_{\mathcal{P}}$, prognostication encoder $R_{\mathcal{C}}$, and demographic encoder $R_{\mathcal{D}}$, as defined in Equations \ref{eq:enc_rp}, \ref{eq:ep} and \ref{eq:enc_rd}, respectively. 
The set $\mathbb{P}=\{P^{\mathcal{D}}, P^\mathcal{C}\}$ denotes the trainable prototype vectors of demographic cohorts and clinical courses, respectively.
Finally, $\mathbb{W}=\{\mathbb{W}^{\mathcal{D}}, \mathbb{W}^{\mathcal{C}}, \Delta \mathbb{W}^{\mathcal{I}}\}$ contains the weight vectors $\mathbb{W}^{\mathcal{D}}$ and $\mathbb{W}^{\mathcal{C}}$, and the interaction contribution matrix $\Delta \mathbb{W}^{\mathcal{I}}$. The full parameter set $\Phi$ is learned by minimizing the following objective function through mini-batch gradient descent:
\begin{equation}
\Phi^* = \arg \min_{\Phi} \text{CE} + \lambda_d^{\mathcal{D}}L_d^{\mathcal{D}} +\lambda_d^{\mathcal{C}}L_d^{\mathcal{C}} + \lambda_s^{\mathcal{C}}L_s^{\mathcal{C}} + \lambda_p^{\mathcal{C}}L_p^{\mathcal{C}} + \lambda^{\mathcal{I}}|\Delta \mathbb{W}^{\mathcal{I}}|,
\end{equation}
where $\text{CE}$ is the standard binary cross-entropy loss computed for the entire dataset $\mathbf{D}$, and the coefficients ($\lambda_d^{\mathcal{D}}, \lambda_d^{\mathcal{C}}, \lambda_s^{\mathcal{C}}, \lambda_p^{\mathcal{C}}$, $\lambda^{\mathcal{I}}$) are hyperparameters used to balance the contribution of five regularization terms explained below. Specifically, 
\begin{equation}
\text{CE} = \sum_{(X^{(u)},y^{(u)})\in \mathbf{D}} y^{(u)}\log\hat{y}^{(u)} + (1-y^{(u)})\log(1-\hat{y}^{(u)})
\end{equation}
where $y^{(u)}$ is the actual mortality status at discharge for ICU admission $u$, and $\hat{y}^{(u)}$ is the predicted mortality status computed from EHR $X^{(u)}$ in accordance with Equation \ref{eq:Final}. 
In terms of model regularizations, $|\Delta \mathbb{W}^{\mathcal{I}}|$ represents the L1-norm of the interaction contribution matrix, while $L_d^{\mathcal{D}}$ and $L_d^{\mathcal{C}}$ implement the diversity penalty that promotes the discovery of diverse demographic cohorts and clinical courses by encouraging their respective prototypes to be distinct from other prototypes \citep{ming2019interpretable}. Formally,
\begin{flalign}
    L_d^{\mathcal{D}} &= \sum_{i=1}^{N_{\mathcal{D}}}\sum_{j=i+1}^{N_{\mathcal{D}}} \max\large(0,\;d^{\mathcal{D}}_{\text{min}}-\lVert p^{\mathcal{D}}_i-p^{\mathcal{D}}_j\rVert_2\large)^2,  \\
    L_d^{\mathcal{C}} &= \sum_{i=1}^{N_{\mathcal{C}}}\sum_{j=i+1}^{N_{\mathcal{C}}} \max\large(0,\;d^{\mathcal{C}}_{\text{min}}-\lVert p^{\mathcal{C}}_i-p^{\mathcal{C}}_j\rVert_2\large)^2,
\end{flalign}
where $d^{\mathcal{D}}_{\text{min}}$ and $d^{\mathcal{C}}_{\text{min}}$ define minimum inter-prototype distances and are both set to 3 in our study.The sparsity penalty $L_s^{\mathcal{C}}$, adapted from \citet{crabbe2021explaining}, limits each physiological embedding vector to be close to only a few clinical course prototypes:
\begin{equation}
    L_s^{\mathcal{C}} = \frac{1}{|U|} \sum^{|U|}_{i=1}\lVert\operatorname{sort}(s^{\mathcal{C}}_i)-\mathbf{r}_{\alpha}\rVert_2^2
\end{equation}
where $|U|$ is the number of clinical records, $\lVert\cdot\rVert_2$ denotes the L2-norm, and $\operatorname{sort}(\cdot)$ arranges the elements of a health state vector in ascending order. The vector $\mathbf{r}_{\alpha}$ consists of $\operatorname{int}\big((1-\alpha) \cdot N_{\mathcal{C}}\big)$ zeros followed by $\operatorname{int}(\alpha \cdot N_{\mathcal{C}})$ ones, where $\alpha \in (0,1)$  specifies the proportion of clinical course prototypes expected to be activated for each physiological embedding.  Here, $\operatorname{int}(\cdot)$ denotes the integer part (i.e., floor operation) of its argument. In our study, $\alpha$ is set to 0.1. Finally, we incorporate the prognostication loss $L_p^{\mathcal{C}}$, as defined in Equation~\ref{eq:Lp}, to embed prognostication awareness into model reasoning, thereby enhancing ProtoDoctor’s predictive performance.

\subsection{Prototype Interpretation}\label{subsec:PI}
Following \citet{ming2019interpretable} and \citet{chen2019looks}, to interpret the learned clinical course prototypes and demographic cohort prototypes, we project each clinical course prototype onto the closest physiological embedding from the training dataset:

\begin{equation}
     p^{\mathcal{C}}_k \leftarrow h^{\mathcal{P}(u^*)}_T \text{ with } u^* = \arg\max_{\forall u \in U} \cos(h^{\mathcal{P}(u)}_T, p^{\mathcal{C}}_k)
\end{equation}
where $h^{\mathcal{P}(u)}_T$ is computed using Equation \ref{eq:enc_rp} for each admission $u$.
In this case, each clinical course prototype can be interpreted as the complete physiological sequence from a specific ICU admission in the input space.

Similarly, each demographic cohort prototype is visualized as the demographic attributes of an ICU admission in the training dataset, which is closest to the prototype in the embedding space:
\begin{equation}
     p^{\mathcal{D}}_m \leftarrow h^{\mathcal{D}(u^*)} \text{ with }  u^*=\arg\max_{\forall u \in U} \cos(h^{\mathcal{D}(u)}, p^{\mathcal{D}}_m)
\end{equation}
where $h^{\mathcal{D}(u)}$ is generated based on Equation \ref{eq:enc_rd}. These projections enable clinically meaningful interpretation of ProtoDoctor’s reasoning process, which aligns with the ICU decision-making practices. Further details on these interpretations are provided in Section~\ref{sec:interpretation}.

\section{Empirical Evaluation}

\subsection{Data Collection and Evaluation Metrics}
To evaluate the predictive performance and interpretability of ProtoDoctor, we utilized the Medical Information Mart for Intensive Care version 3 (MIMIC-III) database~\citep{johnson2016mimic}, which contains deidentified electronic health records (EHRs) from 61,532 ICU admissions recorded between 2001 and 2012 at a major academic medical center in Boston. Following the data preprocessing protocol established by \citet{harutyunyan2019multitask}, we excluded all hospitalizations involving multiple ICU admissions to avoid ambiguity in labeling outcomes, since mortality in these cases could be associated with the entire hospitalization rather than a specific ICU admission. We also removed admissions for patients under 18 years old, due to substantial physiological differences between pediatric and adult populations. After applying these criteria, our final dataset for ICU mortality prediction consists of 21,139 unique ICU admissions. For feature construction, we adopted the variable selection strategies of \citet{harutyunyan2019multitask} and \citet{ma2020concare}, incorporating 20 clinical variables that include both demographic attributes and physiological features. Demographic attributes are obtained following \citet{ma2020concare}, while physiological features are selected according to \citet{harutyunyan2019multitask}. By concatenating the sequences of all physiological features, we obtained a 76-dimensional physiological feature vector $\mathcal{P}_t$ at each hour $t$. The details are provided in Appendix~\ref{appendix:features}. For each ICU admission, we extracted sequences of physiological variables from the first 48 hours in addition to demographic attributes. These sequences were preprocessed according to the protocol of \citet{harutyunyan2019multitask}, including imputation, discretization, and standardization to ensure consistency with established clinical time-series modeling benchmarks. The mortality status at discharge is defined as a binary indicator at the admission level: 1 if the patient died during the ICU admission, and 0 otherwise. Among the 21,139 admissions in the final dataset, 2,797 cases (13.23\%) result in death, indicating a notable class imbalance. To construct the training and evaluation sets, we partitioned the data into training (14,681), validation (3,222), and test (3,236) sets, reflecting an approximate 70\%/15\%/15\% split. To prevent information leakage and ensure the validity of model evaluation, all admissions from a single patient are assigned exclusively to one partition. We assessed the predictive performance of ProtoDoctor by comparing its predictions with ground-truth labels in the test dataset, using two widely adopted metrics in ICU mortality prediction research: the Area Under the Receiver Operating Characteristic Curve (AUROC) and the Area Under the Precision-Recall Curve (AUPRC)~\citep{ma2020concare,ma2023mortality}. AUROC measures the model’s capacity to distinguish between mortal and non-mortal ICU admissions, where a score of 1.0 indicates perfect discrimination and 0.5 represents random guessing. AUPRC assesses the model’s ability to accurately identify mortal ICU admissions, capturing the balance between precision (correctly identified mortal admissions among those predicted as mortal) and recall (correctly identified mortal admissions among actual mortal cases). A score of 1.0 reflects perfect performance, while a value near the mortality rate reflects performance close to random guessing. Given the substantial class imbalance in our ICU mortality prediction dataset (mortality rate: 13.23\%), we placed greater emphasis on AUPRC as it directly evaluates the model’s effectiveness in detecting actual mortal cases, making it more suitable for the ICU context. For each model, we reported the mean and standard deviation of AUROC and AUPRC over five experimental runs to ensure statistical robustness.

\subsection{Benchmark Methods}
We evaluated the performance of our proposed model, ProtoDoctor, against two primary categories of benchmark methods: (1) interpretable DL models designed for healthcare prediction tasks, and (2) recent prototype learning approaches adaptable for interpretable ICU mortality prediction. State-of-the-art interpretable healthcare prediction methods commonly employ DL models with attention mechanisms to provide intrinsic interpretability. These models attribute importance scores to input features through attention weights. Accordingly, we first included ConCare~\citep{ma2020concare} as a benchmark. ConCare is an attention-based ICU mortality prediction framework that achieves state-of-the-art performance and offers interpretability by modeling cross-feature interdependencies.
Then, we also evaluated AICare~\citep{ma2023mortality}, which assigns attention weights to features at each time step. When adapted to ICU data, each attention weight generated by AICare reflects the importance of a clinical feature at a specific hour for ICU mortality prediction. While attention-based models provide interpretability at the feature level, prototype learning methods make it possible to interpret based on representative clinical courses and patient cohorts, mirroring the ICU decision-making practices more closely. Since no prior studies have applied prototype learning methods to ICU mortality prediction, we adapted two representative approaches originally developed for other tasks as benchmark models: ProtoryNet~\citep{hong2023protorynet} and PahNet~\citep{xie2024prototype}. In our context, ProtoryNet models each prototype as a typical static state of physiological features at a specific hour, enabling clinical course-based interpretations through sequences of state-level prototypes. In contrast, PahNet interprets each prototype as a complete physiological sequence, thus directly capturing prototypical clinical courses. To incorporate demographic information without compromising interpretability, we augmented each prototype-based baseline with a demographic branch, utilizing the same demographic encoder and demographic cohort prototypes as employed in ProtoDoctor. Table~\ref{tb:benchmethods} provides an overview of all benchmark methods considered in our evaluation.

\begin{table}[h]
	\caption{Methods Compared in Our Evaluation}
	\begin{center}

		\begin{tabular}{
				L{150pt}
				L{300pt} } 
			\hline
			Method	& \multicolumn{1}{c}{Notes}  \Tstrut\\
			\hline
			ProtoDoctor & Our proposed method  \Tstrut\\
			ConCare \citep{ma2020concare} & Representative attention-based interpretable ICU mortality prediction method  \Tstrut\\
			AICare \citep{ma2023mortality} & Representative attention-based interpretable healthcare prediction method  \\
			ProtoryNet \citep{hong2023protorynet} & State-level Prototype learning method adapted for ICU mortality prediction \\
            PahNet \citep{xie2024prototype} & Sequence-level Prototype learning method adapted for ICU mortality prediction \\
			\hline
		\end{tabular}
	\end{center}
	\label{tb:benchmethods}
\end{table}

Next, we outline the implementation details for all benchmark models. Each model processes both a dynamic physiological sequence and a vector of static demographic attributes as inputs and predicts the mortality status at discharge for each ICU admission. 
We followed the implementation and configuration of ConCare provided by the authors, since it utilizes the same dataset as ProtoDoctor\footnote{The implementation of ConCare is based on the official code: https://github.com/Accountable-Machine-Intelligence/ConCare}. 
For all other baseline models, we performed hyperparameter tuning on the validation set and reported predictive performance on the test set. ProtoDoctor encompasses three specialized encoders. 
The physiological encoder ($R_{\mathcal{P}}$) employs a channel-wise architecture following \citet{harutyunyan2019multitask}, with the sequence of each physiological feature independently processed by a two-layer bidirectional LSTM (8 hidden units per layer). Feature-specific embeddings are concatenated hourly to form a unified temporal representation, which is then processed by time-wise and feature-wise attention layers. Skip connections and a dropout rate of 0.5 are applied to enhance training stability and prevent overfitting. 
The demographic encoder ($R_{\mathcal{D}}$) is a single-layer MLP with 64 hidden units and Tanh activation, generating the embedding vector of demographic attributes. 
The prognostication encoder ($R_{\mathcal{C}}$) operates as a single-layer LSTM whose embedding dimension size equals the number of clinical course prototypes ($N_{\mathcal{C}}$). This encoder predicts the future health state vector based on the sequence of preceding health state vectors to embed prognostication into model reasoning. Other key hyperparameters for ProtoDoctor include the number of clinical course prototypes ($N_{\mathcal{C}} = 50$), the number of demographic prototypes ($N_{\mathcal{D}} = 20$), and regularization coefficients ($\lambda_d^{\mathcal{D}}, \lambda_d^{\mathcal{C}}, \lambda_s^{\mathcal{C}}, \lambda_p^{\mathcal{C}}, \lambda^{\mathcal{I}}$), which were set to $1\times10^{-3}$, $1\times10^{-3}$, $5\times10^{-1}$, $5\times10^{-2}$, and $1\times10^{-3}$, respectively. For AICare~\citep{ma2023mortality}, the key hyperparameters include the number of hidden units in the channel-wise temporal encoder (8), the embedding size in the multi-head attention layer (32), the feed-forward size (256), and the dropout rate (0.3).\footnote{The implementation of AICare is based on the official code: https://github.com/Accountable-Machine-Intelligence/AICare.} For ProtoryNet~\citep{hong2023protorynet}, which was initially developed for text classification, we replaced the original pretrained text embedding module with ProtoDoctor’s physiological encoder $R_{\mathcal{C}}$, set the LSTM hidden size to 128, and used a dropout rate of 0.5. For PahNet~\citep{xie2024prototype}, originally designed for ECG classification with fewer input channels, we increased the Bi-LSTM hidden units to 64 to accommodate the higher dimensionality of physiological data. For both prototype-based baselines, the numbers of clinical course prototypes and demographic prototypes are set to 50 and 20, respectively. All models were implemented in PyTorch and optimized using the Adam optimizer~\citep{kingma2014adam} with a learning rate of $1\times10^{-3}$. The batch size was set to 32 for ProtoDoctor and other prototype-based baselines, while it was 256 for attention-based baselines. Training was conducted for up to 100 epochs, with early stopping based on validation loss and a patience of 10 epochs to prevent overfitting.

\subsection{Evaluation Results}
Table~\ref{Table_Perfeval} presents the performance metrics (mean$\pm$standard deviation) for ProtoDoctor and baseline methods across five independent runs, with statistical significance evaluated via t-tests. Specifically, ProtoDoctor achieves the highest AUPRC (0.5417 $\pm$ 0.0101), significantly outperforming all baselines ($p < 0.10$), while attaining a competitive AUROC that is statistically indistinguishable from the top-performing ConCare model. As noted by prior studies~\citep{harutyunyan2019multitask,ma2020concare}, the observed AUPRC–AUROC gap reflects the inherent class imbalance in ICU mortality prediction, where mortal cases constitute only a small fraction of admissions. While AUROC measures overall discriminative ability between mortal and non-mortal cases, AUPRC focuses on the mortal class, making it more aligned with the primary clinical goal of accurately identifying high-risk (mortal) ICU patients. Accordingly, ProtoDoctor’s superiority in AUPRC underscores its enhanced clinical utility for ICU mortality prediction compared to existing approaches.\footnote{We also noted that the absolute performance values reported here are comparable to, but slightly lower than those in \citet{ma2020concare}, likely due to differences in evaluation protocols, such as the use of bootstrapping in prior work.}

\begin{table}[h]
    \centering
    \caption{Prediction Performance Comparison between ProtoDoctor and Baselines (T=48)}\label{Table_Perfeval}
    \begin{threeparttable}
    \begin{tabular}{ccccc}
    \toprule
    Model  & AUROC & AUPRC\\ \hline
    ProtoDoctor(Ours) & $0.8668\pm 0.0013$ & \textbf{0.5417$\pm$0.0101}\\
    ConCare \citep{ma2020concare}  & \textbf{0.8673$\pm$0.0011} & $0.5095\pm0.0110^{***}$\\
    AICare \citep{ma2023mortality}  & $ 0.8472\pm0.0077^{***}$ & $0.4800\pm0.0119^{***}$ \\ 
    ProtoryNet \citep{hong2023protorynet} & $0.8579\pm0.0057^{**}$ & $0.5214\pm0.0126^{**}$ \\
    PahNet \citep{xie2024prototype} & $0.8491\pm0.0059^{***}$ & $0.4681\pm0.0147^{***}$ \\
    
    \bottomrule
    \end{tabular}
    \begin{tablenotes} \footnotesize
       \item Note: $^{*}p<0.10; ^{**}p<0.05; ^{***}p<0.01$
    \end{tablenotes}
    \end{threeparttable}
\end{table}

To ensure the robustness of our evaluation results, we conducted additional experiments by reducing the observation period $T$ of physiological sequences from 48 to 24 hours, which is also a commonly adopted setting in prior ICU mortality prediction research~\citep{iwase2022prediction}. 
As reported in Table~\ref{tb:robustness}, ProtoDoctor consistently achieves the highest AUPRC and the second-highest AUROC across all baseline comparisons. Specifically, ProtoDoctor outperforms the second-best method, ProtoryNet, by approximately 1.0\% in AUPRC, and achieves even greater performance margins (up to 10.0\%) over other baselines. These statistically significant improvements further validate the robustness of ProtoDoctor’s advantage in predictive performance.
We also note that prototype learning–based methods generally achieve higher AUPRC than non-prototype approaches, corroborating their effectiveness in identifying mortal admissions, particularly when the available clinical information is limited.

\begin{table}[h]
    \centering
    \caption{Prediction Performance Comparison between ProtoDoctor and Baselines (T=24)}
    \begin{threeparttable}
    \begin{tabular}{ccccc}
    \toprule
    Model  & AUROC & AUPRC\\ \hline
    ProtoDoctor(Ours) & $0.8146\pm0.0027 $ & \textbf{0.4000}$\pm$\textbf{0.0062}\\
    ConCare \citep{ma2020concare}  & \textbf{0.8161}$\pm$\textbf{0.0022} & $ 0.3751\pm0.0155^{**}$\\
    AICare \citep{ma2023mortality}  & $0.7950\pm0.0193^{*}$ & $0.3639\pm0.0259^{**}$ \\ 
    ProtoryNet \citep{hong2023protorynet} & 0.8094$\pm$0.0038$^{**}$ & 0.3962$\pm$0.0079 \\
    PahNet \citep{xie2024prototype} & $ 0.8057\pm0.0023^{***}$ & $ 0.3764\pm0.0136^{**}$ \\
    \bottomrule
    \end{tabular}
    \begin{tablenotes} \footnotesize
       \item Note: $^{*}p<0.10; ^{**}p<0.05; ^{***}p<0.01$
    \end{tablenotes}
    \end{threeparttable}
    \label{tb:robustness}
\end{table}

\subsection{Performance Analysis}
Methodologically, ProtoDoctor features two key innovations: (i) the prognostication-aware regularization (PAR) component in the PCCI module and (ii) the D-C interaction (DCI) component in the DHR module. To assess the individual contribution of each component, we conducted ablation studies by sequentially removing these modules and evaluating the impact on predictive performance. First, we evaluated the impact of the DCI component by removing it from ProtoDoctor and referred to the resulting model as ProtoDoctor-D. The performance difference between ProtoDoctor and ProtoDoctor-D reveals the contribution of accounting for the demographic heterogeneity to the performance of ProtoDoctor. Next, we further removed the PAR component from  ProtoDoctor-D and referred to the resulting variant as ProtoDoctor-DP. The performance difference between ProtoDoctor-D and ProtoDoctor-DP uncovers the contribution of prognostication awareness to ProtoDoctor's performance. We compared the performance of ProtoDoctor-DP, ProtoDoctor-D, and ProtoDoctor under the primary evaluation setting ($T=48$). As reported in Table~\ref{Table_ablation}, 
ProtoDoctor outperforms ProtoDoctor-D by 0.64\% in AUROC and 2.01\% in AUPRC, whereas ProtoDoctor-D surpasses ProtoDoctor-DP by 0.02\% and 1.51\% in AUROC and AUPRC, respectively. In summary, both components significantly enhance predictive performance, with particularly notable improvements in AUPRC. These findings confirm that ProtoDoctor’s methodological advances are especially effective in improving the identification of mortal cases, a critical capability given the severe class imbalance in ICU mortality prediction.

\begin{table}[h]
    \centering
    \caption{Ablation Analysis of ProtoDoctor}
    \begin{tabular}{L{80pt} C{80pt} C{80pt} C{80pt} C{80pt}} 
    \toprule
        & PAR & DCI & AUROC & AUPRC \\ \hline
    ProtoDoctor-DP  &  &  &  0.8611$\pm$0.0021  &  0.5231$\pm$0.0085          \\
    ProtoDoctor-D   & $\surd$ &  & 0.8613$\pm$0.0020 & 0.5310$\pm$0.0025     \\
    ProtoDoctor     & $\surd$ & $\surd$ & \textbf{0.8668$\pm$0.0013}  & \textbf{0.5417$\pm$0.0101} \\
    \bottomrule
    \end{tabular}
    \label{Table_ablation}
\end{table}

\subsection{Interpretability of ProtoDoctor}\label{sec:interpretation}
\begin{figure}[h]
    \centering
    \includegraphics[width=0.8\linewidth]{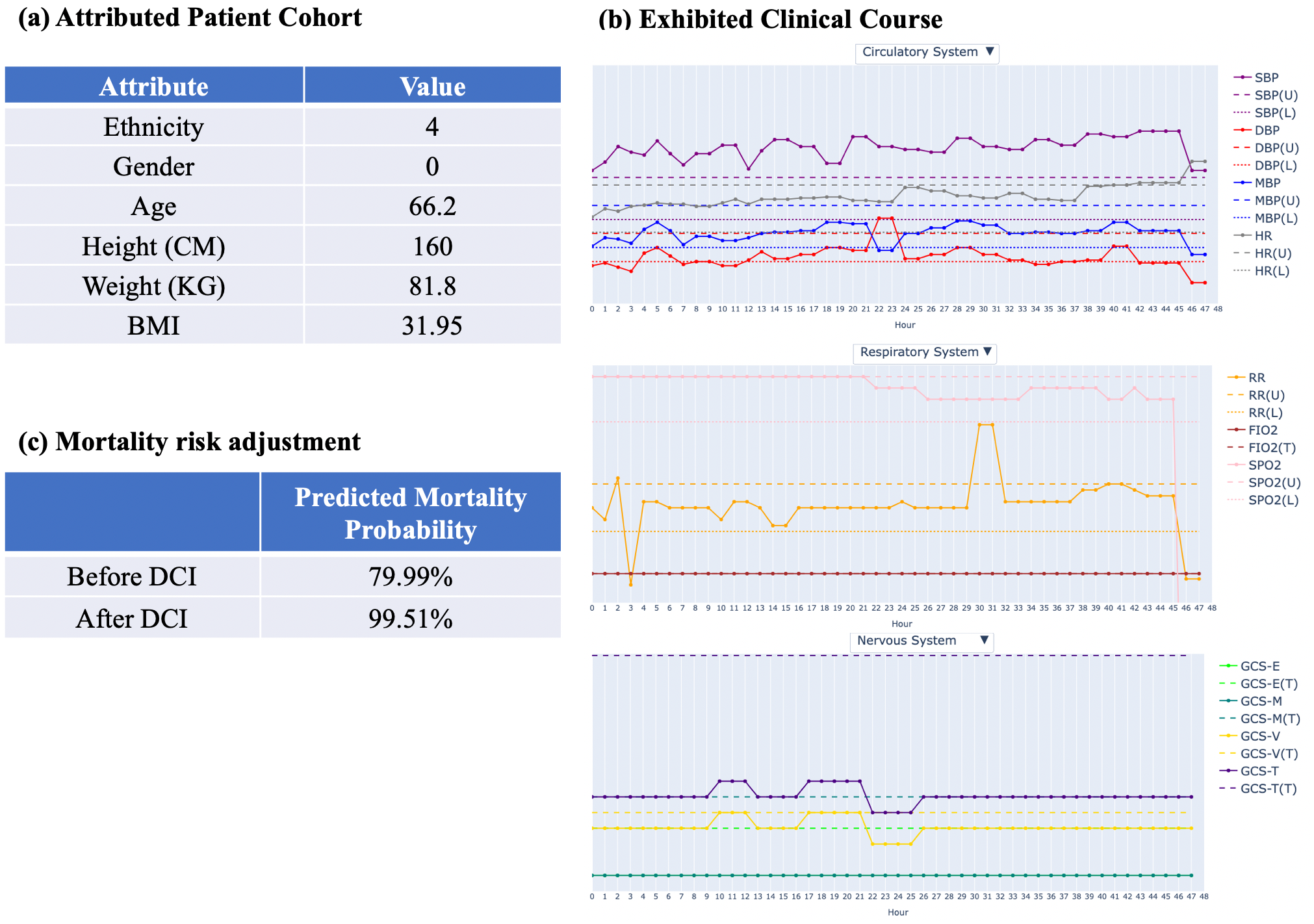}
    \caption{Interpretation of a mortal ICU patient}
    \label{fig:pos_case}
\end{figure}

\begin{figure}[h]
    \centering
    \includegraphics[width=0.8\linewidth]{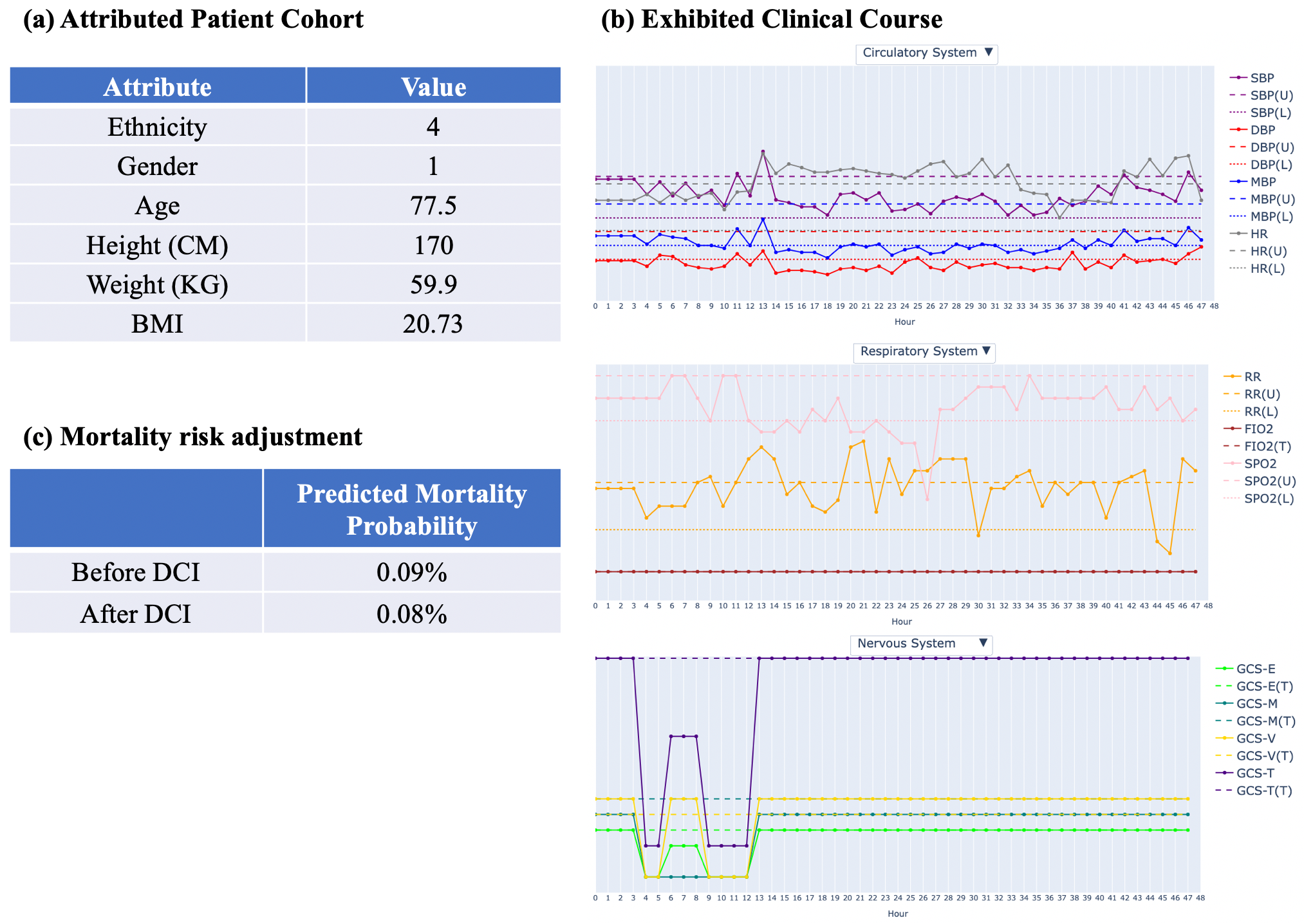}
    \caption{Interpretation of a surviving ICU patient}
    \label{fig:neg_case}
\end{figure}

Beyond outstanding predictive performance compared with state-of-the-art baselines, ProtoDoctor offers clinically meaningful model interpretability for ICU mortality prediction by identifying the exhibited clinical courses, attributing patients to demographic cohorts, and quantifying personalized mortality risk adjustments. To illustrate the model’s reasoning process, we presented two representative cases, one involving a patient who died and one who survived, as shown in Figures \ref{fig:pos_case} and \ref{fig:neg_case}. Each figure consists of three panels: the attributed patient cohort (Panel a), the most aligned clinical course prototype (Panel b), and the mortality risks before and after incorporating the personalized mortality risk adjustment via the DCI component (Panel c). In each figure, Panel (a) summarizes the demographic attributes of the attributed cohort for each patient, including ethnicity, gender, age, height, weight, and body mass index (BMI). Panel (b) visualizes the clinical course prototype most aligned with the patient’s physiological sequence as sequences of key physiological variables. These variables are organized by the circulatory, respiratory, and neurological systems, respectively (from top to bottom), with their hourly measurements appearing as solid lines. For variables with established normal ranges, the upper and lower bounds are shown using dashed and dotted lines, respectively. These reference lines are labeled with corresponding variable abbreviations followed by $(\text{U})$ for upper bound or $(\text{L})$ for lower bound. For example, ``SBP$(\text{U})$" and ``SBP$(\text{L})$" symbolize the upper and lower bounds of systolic blood pressure. For variables considered abnormal when exceeding a defined threshold, a horizontal dashed line indicates the threshold and is marked with the variable abbreviation followed by $(\text{T})$. For instance,  GCS-T$(\text{T})$ denotes the threshold value of the Glasgow Coma Scale (GCS) total score, with values below 15 suggesting neurological impairment according to established medical guidelines~\citep{teasdale1974assessment}. Panel (c) presents the predicted mortality probabilities before and after considering personalized mortality risk adjustments, demonstrating how patient-specific cohort information influences the predicted mortality status.

ProtoDoctor predicts the patient in Figure~\ref{fig:pos_case} as mortal based on three main reasons.
First, as shown in Panel (a), the patient belongs to a demographic cohort characterized by obesity, which has been widely recognized as a risk factor linked to elevated ICU mortality~\citep{hogue2009impact}. Second, the patient’s physiological sequence closely matches a clinical course prototype that reflects progressive multi-organ failure, as illustrated in Panel (b). A sustained reduction in diastolic and mean blood pressures below established normal thresholds indicates worsening hypotension, a condition strongly associated with elevated mortality in critical care settings~\citep{schuurmans2024hypotension}. At the same time, key respiratory indicators such as respiratory rate and oxygen saturation demonstrate deteriorating trends consistent with respiratory failure~\citep{garrido2018respiratory}. Neurological evaluation shows persistently low GCS scores, suggesting a prolonged state of deep coma~\citep{teasdale1974assessment}.  Taken together, these patterns collectively reflect severe multi-organ dysfunction and suggest that the patient faces a very high risk of in-hospital mortality. Finally, cohort-specific mortality risk adjustment elevated the predicted mortality probability from 79.99\% to 99.51\%,  as reported in Panel (c). This adjustment further emphasizes the heightened mortality risk associated with the patient’s attributed cohort.

Similarly, ProtoDoctor predicts survival for the patient in Figure~\ref{fig:neg_case} according to three key factors. First, as shown in Panel (a), the patient is assigned to a demographic cohort with a normal body mass index (BMI) of 20.73, a value generally associated with lower ICU mortality risk. Second, the patient’s physiological sequence closely resembles a clinical course prototype that reflects stabilization and recovery, as illustrated in Panel (b): Circulatory variables exhibit some fluctuation but largely remain within established normal ranges, indicating preserved cardiovascular function. Respiratory measurements also stay near normal levels throughout the observation period.  Although the GCS scores initially drop to an abnormal level, they gradually return to normal, signifying neurological recovery.  This prototype aligns with a typical clinical course associated with patient survival. Third, the cohort-specific mortality risk adjustment further confirms the survival prediction. The estimated mortality probability remains low before and after personalized adjustment, as reported in Panel (c). This result demonstrates the model’s strong confidence in its survival assessment.

\begin{figure}
    \centering
    \includegraphics[width=0.5\linewidth]{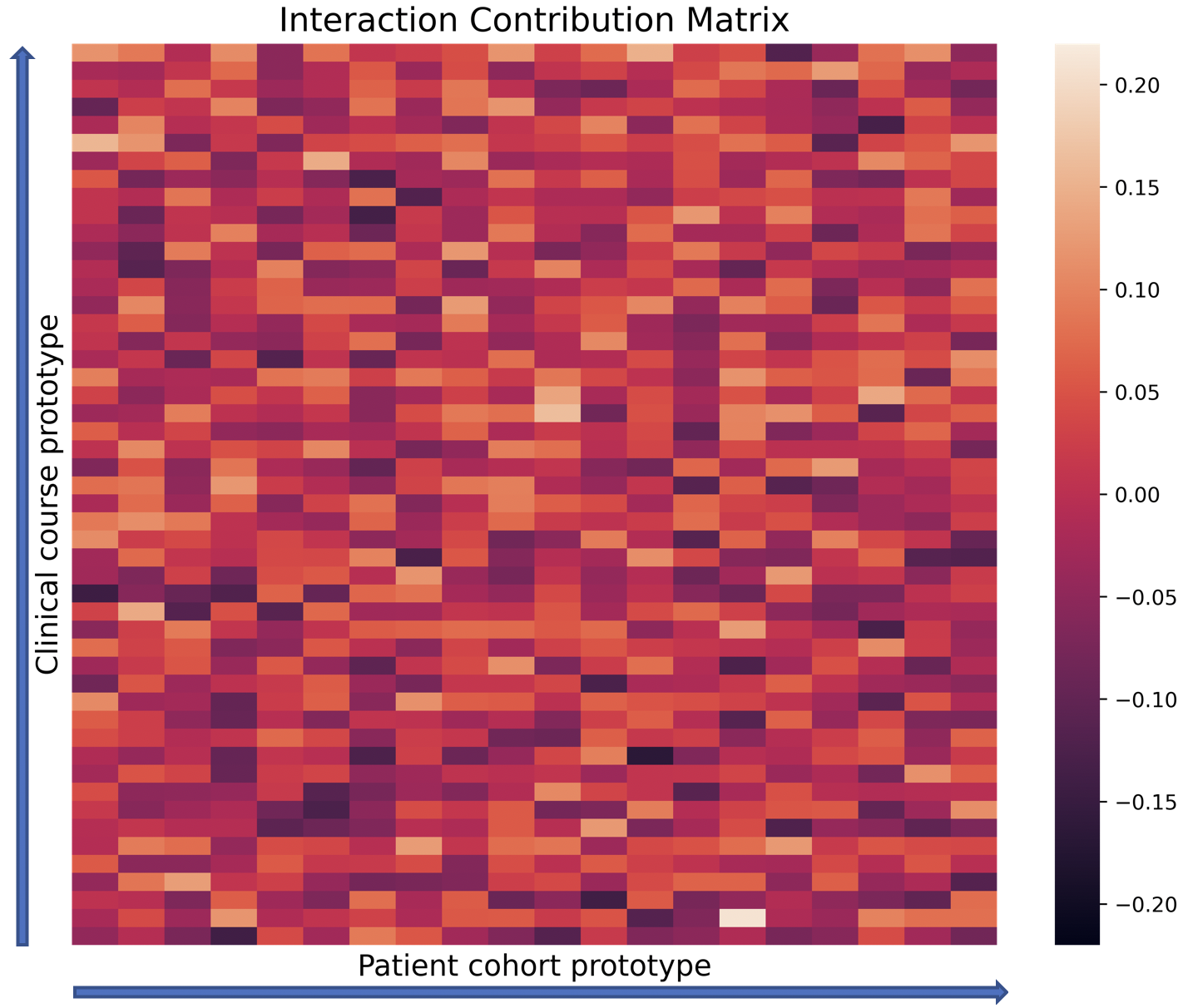}
    \caption{Interaction contribution matrix visualization}
    \label{fig:itr_mat}
\end{figure}

Furthermore, Figure~\ref{fig:itr_mat} displays the learned interaction contribution matrix, $\Delta \mathbb{W}^{\mathcal{I}}$. In this matrix, prototypes of demographic cohorts and clinical courses are arranged in ascending order according to their associated mortality weights derived from $\mathbb{W}^{\mathcal{D}}$ and $\mathbb{W}^{\mathcal{C}}$. The direction of sorting is indicated by the vertical and horizontal arrows along the axes. A positive contribution weight $\Delta w^{\mathcal{I}}_{k,m}$ indicates an elevated mortality risk adjustment of the clinical course $k$ for demographic cohort $m$. The diversity of the interaction weights between demographic cohorts and clinical courses reveals ProtoDoctor's capability to capture heterogeneous mortality risk adjustments of each clinical course associated with different demographic cohorts. Some interaction patterns show potential for clinically meaningful interpretations that may assist ICU doctors in tailoring risk assessment and decision-making. These implications and their alignment with expert judgments are further explored in the Human Evaluation section.

\subsection{Human Evaluation}
To rigorously evaluate the interpretability of ProtoDoctor, we conducted a structured user study with a panel of 40 ICU experts (8 PhDs and 32 MDs) from a leading Emergency Department in China. These participants have an average of 15.5 years of ICU experience. Notably, our expert panel is substantially larger than those in previous healthcare IS studies, where the number of experts rarely exceeded eight~\citep{kim2023rolex, guo2024explainable, xie2025care}, and even in clinical diagnostic research, where the largest reported panel was nine~\citep{bertens2013use}. This significantly larger panel strengthens the generalizability and robustness of our evaluation. Each expert independently participates in a 20-minute interview comprising two structured surveys designed to evaluate ProtoDoctor's interpretability in terms of both the clinical reasoning process and the generated clinical statements. 

\begin{figure}
    \centering
    \includegraphics[width=0.75\linewidth]{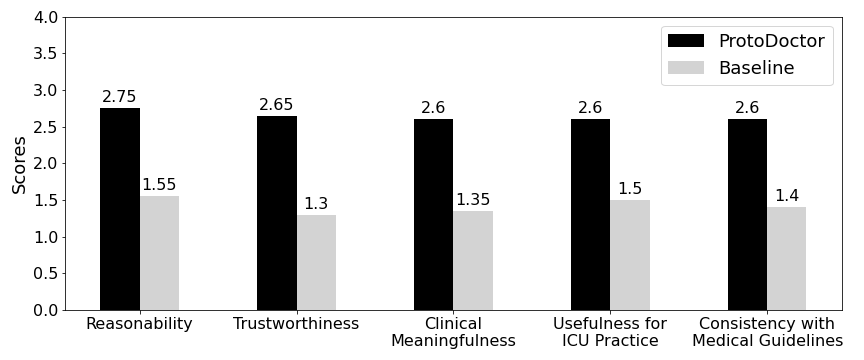}
    \caption{Evaluation on the Interpretation of Reasoning Process (Mean in Each Group)}
    \label{fig:HE}
\end{figure}

In the first survey, experts are randomly assigned to review model-generated interpretations from either ProtoDoctor (Figure~\ref{fig:itp_pdoctor}) or ProtoryNet (Figure~\ref{fig:itp_protory}), the best-performing prototype-based baseline. Each expert assesses two ICU admission cases, one correctly predicted as death and one as survival, and then rates the model’s interpretability across five dimensions: reasonability, trustworthiness, clinical meaningfulness, usefulness for ICU practice, and consistency with medical guidelines. The five questions, adapted from \citet{xie2025care}, are:
(1) How reasonable do you think the model’s reasoning process is?
(2) How trustworthy do you think the model’s reasoning process is?
(3) How clinically meaningful do you think the model’s interpretation is for ICU practices?
(4) How useful do you think the model’s summarized typical demographic cohorts and clinical courses are for  ICU mortality prediction?
(5) How consistent do you think the model’s summarized typical clinical courses are with established medical guidelines? 
Ratings are collected using a 5-point Likert scale ranging from 0 (``not at all") to 4 (``extremely"). As shown in Figure~\ref{fig:HE}, ProtoDoctor consistently achieves higher mean scores than the baseline across all criteria, for example, 2.75 versus 1.55 for reasonability, and 2.65 versus 1.30 for trustworthiness. Qualitative feedback from ICU experts further highlights a key limitation of the baseline: relying solely on sequences of typical clinical states is viewed as insufficient for comprehensive mortality assessment. In contrast, ProtoDoctor’s interpretations, incorporating entire clinical courses and personalized mortality risk adjustments, are considered more aligned with the real-world ICU decision-making practices.

The second survey evaluates the clinical plausibility of statements derived from ProtoDoctor’s reasoning, specifically those describing how the adjusted mortality risk of a given clinical course varies across demographic cohorts. Participants are randomly assigned to one of two groups: one reviews two original statements generated by ProtoDoctor, and the other reviews their logical opposites. Following the evaluation format of \citet{guo2024explainable}, experts rate each statement on a 5-point Likert scale (1 = ``strongly disagree" to 5 = ``strongly agree"). The complete statements and their corresponding mean scores are presented in Table~\ref{table_survey}. Results show significantly higher agreement with ProtoDoctor’s statements (e.g., 3.65 versus 2.25). Furthermore, qualitative feedback from experts supports the plausibility of the statements. For instance, the first statement is consistent with existing clinical guidelines regarding the interaction between obesity and hypotension in critical care, where hypotension signals hemodynamic instability and, when coupled with obesity, is associated with elevated mortality risk~\citep{dickerson2022obesity}. Similarly, the second statement reflects the well-documented observation that elderly patients face higher stroke-related mortality~\citep{soto2020age}.

\section{Conclusion}

\subsection{Summary and Contributions}
Accurate ICU mortality prediction is critical given ICU patients' severe conditions and the resource-intensive nature of critical care. The high-stakes environment further necessitates model interpretability to support clinical trust and ensure regulatory compliance. To enhance both predictive performance and interpretability, it is suggested that an ICU mortality prediction model should provide intrinsic interpretations and integrate three critical elements of the ICU decision-making practices into its reasoning process: clinical course identification, demographic heterogeneity, and prognostication awareness. However, while some existing interpretable healthcare models provide intrinsic interpretability, they typically only address demographic heterogeneity. Prototype learning approaches show promise for delivering intrinsic interpretations while incorporating clinical course identification; however, how to achieve demographic heterogeneity and prognostication awareness in prototype learning frameworks remains under-investigated. To address these limitations, we propose ProtoDoctor, a novel interpretable framework for ICU mortality prediction. Specifically, it generates intrinsic interpretations via prototype learning and explicitly incorporates all three essential elements using two innovative components: the prototypical clinical course identification module and the demographic heterogeneity recognition module. Extensive empirical evaluations demonstrate that ProtoDoctor not only outperforms leading interpretable healthcare prediction models and state-of-the-art prototype learning methods in predictive performance but also offers intrinsic interpretations with enhanced interpretability in terms of various dimensions.

Our work belongs to the computational design science (CDS) research genre within the Information Systems (IS) field~\citep{padmanabhan2022machine}. In general, CDS research focuses on developing novel computational methods to solve important business and societal problems, thus making methodological contributions to the literature. 
Specifically, this study addresses a critical healthcare task of ICU mortality prediction by proposing ProtoDoctor, a novel intrinsically interpretable prediction model based on prototype learning. To enable intrinsic interpretability and incorporate critical elements in the ICU decision-making practices, the key novelties of ProtoDoctor lie in the prognostic clinical course identification module and the demographic heterogeneity recognition module, which together form the core methodological contributions of this study.
Additionally, our study also contributes to research in the Health Information Technology (HIT) field by introducing a novel interpretative framework that effectively solves a fundamental problem in the high-stakes ICU settings.

\subsection{Research and Managerial Implications}
Our study offers several important research implications.
First, we advance prototype learning by introducing a modeling approach that operates on the multimodal data structure (i.e., physiological sequence and demographic attributes within EHRs) and captures cross-modal prototype interactions. This addresses key limitations of existing prototype learning models, which typically focus on a single data source. Second, we emphasize the critical role of domain knowledge in designing computational artifacts in CDS research. ProtoDoctor’s architecture is explicitly informed by the ICU decision-making practices, which not only improves predictive performance but also enhances interpretability. This supports and extends longstanding CDS literature emphasizing the value of integrating domain expertise into IT artifact development~\citep{gregor2013positioning,abbasi2018text,he2019mobile}, especially in high-stakes healthcare contexts~\citep{wang2021predicting,zhang2025ketch}.

ProtoDoctor also brings substantial benefits to various healthcare stakeholders in high-stakes ICU settings~\citep{imrie2023multiple}. For patients and their families, interpretability grounded in ICU clinical decision-making practices fosters a clearer understanding of mortality risk assessments, enabling more informed treatment decisions. By referring to historical ICU admissions with similar reasoning logics, patients and families can better comprehend potential outcomes and associated burdens, supporting choices that align with their preferences and values. For ICU doctors, ProtoDoctor builds trust in AI-driven predictions by aligning interpretability with clinical practice. Doctors can rapidly assess whether the model’s rationale is consistent with their expertise. As shown in our human evaluation, ICU doctors found ProtoDoctor’s explanations more clinically meaningful than those of baseline models, supporting safer and more effective care, and promoting seamless AI adoption in ICU workflows. For healthcare regulators and administrators, the transparency inherent in ProtoDoctor provides clear insight into the model’s decision-making logic, thereby fostering clinical trust, enabling rigorous validation, and facilitating its practical deployment in ICU settings. The model interpretability in the form of the desired paradigm is also essential for meeting regulatory standards such as GDPR~\citep{voigt2017eu}, ensuring accountability in critical care decisions, and advancing stakeholder confidence in health information technology adoption.

\subsection{Limitations and future Work}
Our study has limitations and can be extended in multiple ways. Firstly, although ProtoDoctor currently uses only physiological sequences and demographic attributes from a single ICU admission, recent studies show that incorporating additional modalities, such as clinical notes, laboratory results, and medical images, can enhance predictive accuracy in ICU settings~\citep{yang2021multimodal,chen2024multi}. Future work could extend ProtoDoctor to integrate these diverse data sources to enhance predictive performance and provide more comprehensive, clinically meaningful interpretations. Secondly, while ProtoDoctor models clinical courses by jointly capturing temporal patterns across all physiological features, ICU doctors often focus on the co-evolution of specific subsets of critical variables. For example, Fraction of Inspired Oxygen (FiO\textsubscript{2}), Peripheral Capillary Oxygen Saturation (SpO\textsubscript{2}), and respiratory rate (RR) are frequently monitored together, as their combined trends provide key insights into respiratory function and oxygenation~\citep{roca2019index}. Future work could enhance interpretability by explicitly identifying and modeling the temporal dynamics of such clinically meaningful feature subsets. Finally, because ProtoDoctor’s interpretation paradigm closely aligns with the ICU decision-making practices, it holds significant potential for broader clinical applications. Future work could adapt this approach to other high-impact prediction tasks in critical care, such as length-of-stay estimation~\citep{guo2024explainable}, ICU readmission prediction~\citep{xie2021readmission}, and decompensation detection~\citep{harutyunyan2019multitask}.



\bibliographystyle{informs2014} 
\bibliography{related-works} 




\clearpage
\begin{appendices}

\setcounter{figure}{0}
\setcounter{table}{0}
\setcounter{section}{0}
\setcounter{equation}{0}
\setcounter{page}{1}

\renewcommand{\thetable}{A\arabic{table}}
\renewcommand{\thefigure}{A\arabic{figure}}
\renewcommand{\theequation}{A\arabic{equation}}
\renewcommand{\thepage}{EC\arabic{page}}

\section{Description of Clinical Variables}\label{appendix:features}
Table~\ref{table:feature} provides descriptions of input clinical variables and the output label. Following the preprocessing steps of ~\citet{harutyunyan2019multitask}, we first impute missing values in each physiological time series using the most recent observation if available, or a pre-specified ``normal" value otherwise. Normal values are provided in ~\citet{harutyunyan2019multitask}. A binary mask is then added for each variable at every hour to indicate whether the value is observed or imputed. Categorical and binary variables are one-hot encoded, while numeric inputs are standardized by subtracting the mean and dividing by the standard deviation, computed per variable after imputation.
After preprocessing, each physiological variable $i$ is represented by two sequences: $\{\mu_t^{(i)}\}_{t=1}^T$, where $\mu_t^{(i)}$ is a binary mask indicating observation status at hour $t$, and $\{c_t^{(i)}\}_{t=1}^T$, the sequence of observed or imputed values. By concatenating these pairs across all physiological variables, we finally yield a 76-dimensional physiological feature vector per hour.

\begin{center}
\small
\begin{longtable}{P{2.1cm} | P{2.1cm} | P{11.0cm}}
\caption{Description of Clinical variables and Output Variable}
\label{table:feature} \\
\toprule
 Type & Variable name & Description and unit of measure  \\
\hline
\endfirsthead
\hline
 Type & Variable name & Description and unit of measure  \\
\hline
\endhead

Output label & \textit{Mortality status} & Binary (1=the patient is mortal at discharge) \\
\midrule
\multirow{5}{*}{\makecell[c]{Demographic \\ attributes}} & \textit{Ethnicity} & Categorical (0-4, de-identified ethnicity ID) \\
 & \textit{GenderF} & Binary (1=patient is female) \\
 & \textit{Age} & Patient age in years \\
 & \textit{Height} & Patient height in centimeters (CM) \\
 & \textit{Weight} & Patient weight in kilograms (KG) \\
\hline
\multirow{11}{*}{\makecell[c]{Physiological \\ variables}} & \textit{CRR} & Binary (1=Capillary refill rate of the patient is abnormal) \\
 & \textit{DBP} & Diastolic blood arterial pressure in millimeters of mercury (mmHg) \\
 & \textit{MBP} & Mean blood arterial pressure in millimeters of mercury (mmHg)\\
 & \textit{SBP} & Systolic blood arterial pressure in millimeters of mercury (mmHg)\\
 & \textit{FiO$_2$} & Fraction inspired oxygen (float) \\
 & \textit{GCS-E} & Eye response score in Glasgow Coma Scale (GCS) (1–4) \\
 & \textit{GCS-M} & Motor response score in GCS (1–6) \\
 & \textit{GCS-V} & Verbal response score in GCS (1–5) \\
 & \textit{GCS-T} & Total GCS score (3–15) \\
 & \textit{Glucose} & Blood glucose in milligrams per deciliter (mg/dL) \\
 & \textit{HR} & Heart rate in beats per minute \\
 & \textit{SPO$_2$} & Oxygen saturation in percentage \\
 & \textit{RR} & Respiration rate in breaths per minute \\
 & \textit{Temperature} & Body temperature in Celsius \\
 & \textit{pH} & Body pH (float) \\
\bottomrule
\end{longtable}

\end{center}

\setcounter{figure}{0}
\setcounter{table}{0}
\setcounter{equation}{0}

\renewcommand{\thetable}{B\arabic{table}}
\renewcommand{\thefigure}{B\arabic{figure}}
\renewcommand{\theequation}{B\arabic{equation}}

\section{Results of Human Evaluation} \label{appendix:survey}
\begin{figure}[h]
    \centering
    \includegraphics[width=\linewidth]{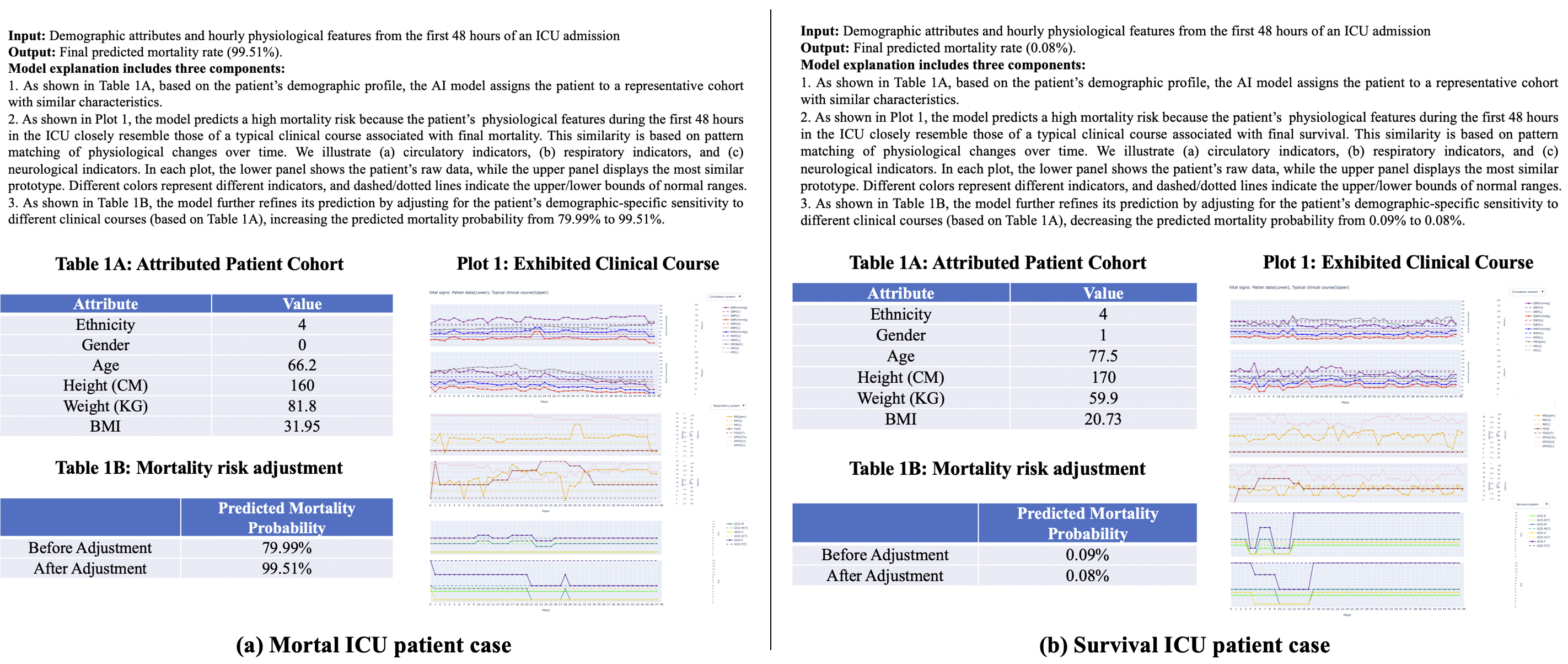}
    \caption{Human evaluation on interpretations of a mortal and a survival ICU patient generated by ProtoDoctor}
    \label{fig:itp_pdoctor}
\end{figure}

\begin{figure}[h]
    \centering
    \includegraphics[width=\linewidth]{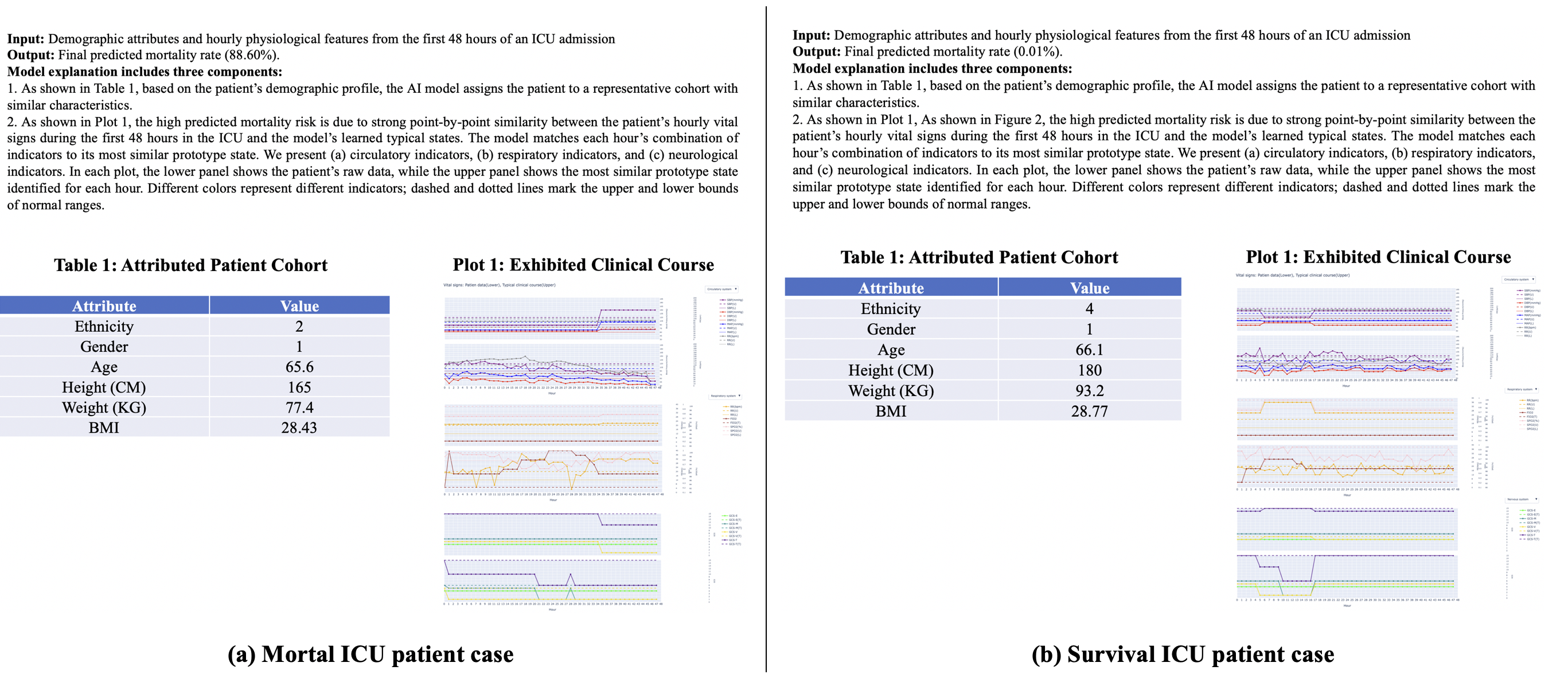}
    \caption{Human evaluation on interpretations of a mortal and a survival ICU patient generated by ProtoryNet}
    \label{fig:itp_protory}
\end{figure}

The survey presented for human evaluation on interpretations generated by ProtoDoctor for a mortal ICU patient and a survival ICU patient is shown in Figure~\ref{fig:itp_pdoctor}. 
The survey presented for human evaluation on interpretations generated by ProtoryNet for a mortal ICU patient and a survival ICU patient is shown in Figure~\ref{fig:itp_protory}.
The results of the survey about medical implications generated from ProtoDoctor are shown in Table~\ref{table_survey}.

\begin{table}[htbp]
\caption{Mean responses to different statements regarding ICU mortality predictions}
\label{table_survey}
\centering
\begin{tabular}{p{12cm} c}
\toprule
\textbf{Statements} & \textbf{Mean Score} \\
& \textbf{(Likert Scale)} \\
\midrule
\multicolumn{2}{l}{\textbf{Statements derived from the clinical implications generated by ProtoDoctor.}} \\
1. Among patients who consistently present with preserved consciousness, low blood pressure, rapid respiration, and high oxygen supplementation requirements, those with a high body mass index (BMI) are predicted to have a \textbf{higher} mortality risk compared to those with a normal BMI. & 3.65 \\
2. Among patients who consistently present with deep coma, elevated systolic blood pressure, and reduced diastolic blood pressure, those of older age are predicted to have a \textbf{higher} mortality risk compared to younger patients. & 4.15 \\
\midrule
\multicolumn{2}{l}{\textbf{Statements that contrast with the implications generated by ProtoDoctor.}} \\
3. Among patients who consistently present with preserved consciousness, low blood pressure, rapid respiration, and high oxygen supplementation requirements, those with a high body mass index (BMI) are predicted to have a \textbf{lower} mortality risk compared to those with a normal BMI. & 2.25 \\
4. Among patients who consistently present with deep coma, elevated systolic blood pressure, and reduced diastolic blood pressure, those of older age are predicted to have a \textbf{lower} mortality risk compared to younger patients. & 1.90 \\
\bottomrule
\end{tabular}
\end{table}

\end{appendices}

\end{document}